\documentclass[	DIV=calc,%
							paper=letter,%
							fontsize=11pt,%
							twocolumn]{scrartcl}	 					

\usepackage[english]{babel}										
\usepackage[protrusion=true,expansion=true]{microtype}				
\usepackage{amsmath,amsfonts,amsthm}					
\usepackage[pdftex]{graphicx}									
\usepackage[svgnames]{xcolor}									
\usepackage[hang, small,labelfont=bf,up,textfont=it,up]{caption}	
\usepackage{epstopdf}												
\usepackage{subfig}													
\usepackage{booktabs}												
\usepackage{fix-cm}													

\usepackage{times}
\usepackage{amsmath}
\usepackage{epsfig}
\usepackage{pst-node}
\usepackage[utf8]{inputenc}
\usepackage{float}
\usepackage{wrapfig}

\usepackage{graphicx}
\usepackage{epsfig}
\usepackage{amsmath,amssymb}
\usepackage{url}

\usepackage{sectsty}													
\allsectionsfont{
	\usefont{OT1}{phv}{b}{n}
	}

\sectionfont{
	\usefont{OT1}{phv}{b}{n}
	}

\usepackage{fancyhdr}												
\usepackage{lastpage}	

\usepackage{lettrine}

\usepackage{titling}															

\newcommand{\HorRule}{\color{DarkGoldenrod}
									  	\rule{\linewidth}{1pt}%
										}

\pretitle{\vspace{-30pt} \begin{flushleft} \HorRule 
				\fontsize{50}{50} \usefont{OT1}{phv}{b}{n} \color{Black} \selectfont 
				}
\title{This is the title of the template article}					
\posttitle{\par\end{flushleft}\vskip 0.5em}

\preauthor{\begin{flushleft}
					\large \lineskip 0.5em \usefont{OT1}{phv}{b}{sl} \color{Black}}
\author{Firstname Lastname, }											
\postauthor{\footnotesize \usefont{OT1}{phv}{m}{sl} \color{Black} 
					Computer Science and Artificial Intelligence Laboratory (CSAIL), Massachusetts Institute of Technology								
					\par\end{flushleft}\HorRule}

\date{}																				

\begin{document} 
\title{Notes on image annotation}

\author{Adela Barriuso, Antonio Torralba\\
}

\maketitle

\abstract{\em
We are under the illusion that seeing is effortless, but frequently the visual system is lazy and makes us believe that we understand something when in fact we don't. Labeling a picture forces us to become aware of the difficulties underlying scene understanding. Suddenly, the act of seeing is not effortless anymore. We have to make an effort in order to understand parts of the picture that we neglected at first glance. 

In this report, an expert image annotator relates her experience on segmenting and labeling tens of thousands of images. During this process, the notes she took try to highlight the difficulties encountered, the solutions adopted, and the decisions made in order to get a consistent set of annotations. Those annotations constitute the SUN database~\cite{xiao10}. 
}

\section{Forward by Antonio Torralba}

\begin{figure}
\centering{
\includegraphics[width=0.9\linewidth]{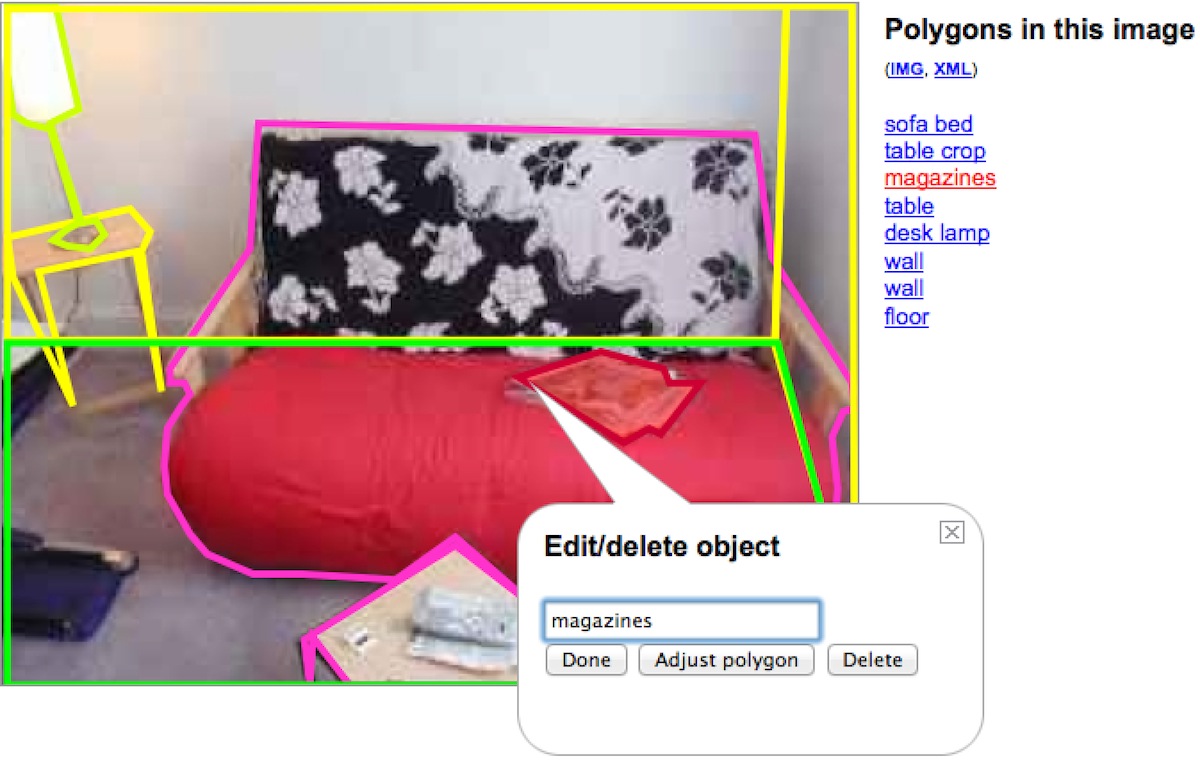}}
\caption{Example of annotated image using the LabelMe image annotation tool.}
\label{fig:tool}
\end{figure}

\begin{figure*}
\centering{
\includegraphics[width=0.45\textwidth]{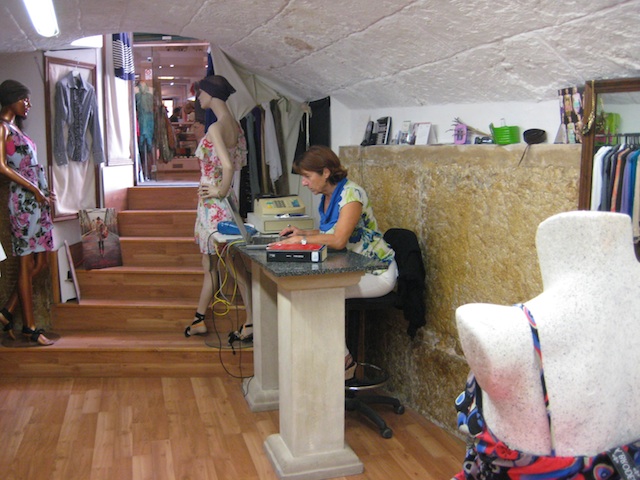}
~~
\includegraphics[width=0.45\textwidth]{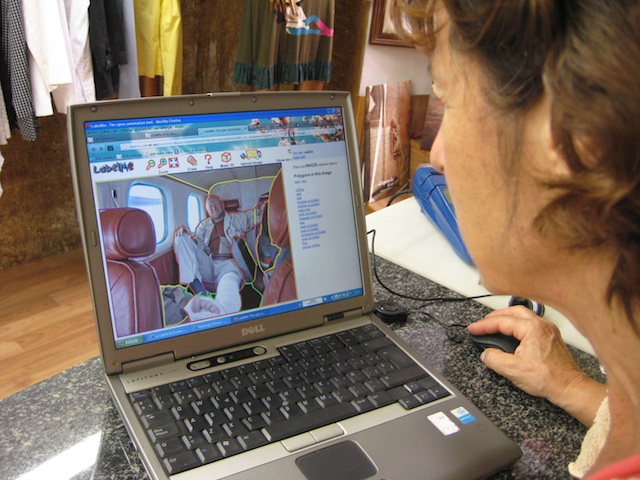}
}
\caption{The image annotation context. All the labeling was done inside a clothing shop named {\em Transparencia} in the heart of Palma de Mallorca, Spain.}
\label{fig:shop}
\end{figure*}

Online games \cite{esp}, Amazon Mechanical Turk \cite{Sorokin08}, crowd-sourcing and a variety of image annotation tools \cite{labelme,vatic} have changed the way data is collected for computer vision research. It would be common to find a student frantically labeling images before a deadline, in order to build up a dataset that would nevertheless be too small to conclude anything reliable  \cite{student}. Those days seem behind us (or are they?). With the prevalence of crowd-sourcing tools, datasets are becoming larger and more ambitious. 

Despite new crowd-sourcing tools allowing the creation of large datasets, it remains important to do some labeling oneself. Labeling images is a good exercise for gaining intuition about possible representations and the limits of the task we are trying to solve. Labeling forces us to clearly think about naming and categorization issues, how to represent occluded objects, how to deal with parts of the image that are unrecognizable, when context becomes important for recognition, what is the effect of our prior knowledge and expertise about a particular visual domain,  and how do we deal with clearly visible objects whose name or function is unknown to us. Where does the identity of an object come from? Does the identity of an object come from its features or from the surrounding objects and our knowledge of typical contextual arrangements? 

It is often said that vision is effortless, but frequently the visual system is lazy and makes us believe that we understand something when in fact we don't. In occasions we find ourselves among objects whose names and even functions we may not know but we do not seem to be bothered by this semantic blindness. However, this changes when we are labeling images as we are forced to segment and name all the objects. Suddenly, we are forced to see where our semantic blind-spot is. We become aware of gaps in our visual understanding of what is around us.




This paper contains the notes written by Adela Barriuso describing her experience while using the LabelMe annotation tool \cite{russell08}. Since 2006 she has been frequently using LabelMe. She has no training in computer vision. In 2007 she started to use LabelMe to systematically annotate the SUN database \cite{xiao10}.  The goal was to build a large database of images with all the objects within each image segmented and named.

Figure~\ref{fig:tool} shows a snapshot of the online annotation tool and one annotated image from the SUN database. Annotating an object requires outlining the object boundary with a polygon and introducing a name as it is shown in Figure~\ref{fig:tool}. A correct annotation should provide a boundary that follows the object outline as accurately as possible (ideally one should be able to recognize the object just by looking at the boundary) and a name that is consistent across different images (e.g. using always the same words to describe objects of the same category and avoiding using synonyms for the same concepts). Defining a labeling protocol is relatively easy when the labeling task is restricted to a few categories, however it becomes challenging when there is not a fix set of categories. As the goal is to label all the objects within each image, the list of categories grows unbounded. Many object classes appear only a few times across the entire collection of images. However, not even those rare object categories can be ignored as they might be an important element for the interpretation of the scene. Labeling in these conditions becomes difficult as it is important to keep a list of all the object classes in order to use a consistent set of terms across the entire database avoiding synonyms. Despite the annotator best efforts, the process is not free of noise.

Since she started working with LabelMe, she has labeled more than 250,000 objects. Labeling more than 250,000 objects gives you a different perspective on the act of seeing. After a full day of labeling images, when you walk on the street or drive back home, you see the world in a different way. You see polygons outlining objects, you start thinking about what they are, and you are especially bothered by occlusions. 
In the rest of the paper I have translated the document written by Adela Barriuso from Spanish to English. I have respected as much as possible the wording and structure of the original document with the exception that I have added subsection headers to try to separate the descriptions to make it easier to read, and I have also added a short conclusion at the end.  




\section{Notes from an annotator by Adela Barriuso}
 
I work in a small clothing shop (Fig.~\ref{fig:shop}). The shop is open from 10am to 8pm with only a short break at 2pm. Despite the long working hours I have a lot of free time. As I am the owner of the shop, I can do whatever I want during that time. I am always ready for the clients, however, in such a long day there are many hours of inactivity. I used to read a lot and books passed by my hands a great speed. I was starting to lose the pleasure that one feels when reading a good book. For this reason, when I started working with LabelMe it was very satisfying to know that I was doing something that had some scientific value and that it could be of use for somebody in the future. 

\subsection{Starting a picture}


To start a new picture is like starting a new challenge. The first thing I do is write down the number of objects that appear in the LabelMe counter to know how many new objects I annotated by the end of the day. It might seem silly, but I always try to surpass myself. 

First I look at the picture to understand it well and to measure the degree of difficulty that it will have. If it is an open space, I start annotating the large surfaces first. Generally I start with the sky and then I continue adding everything else. For enclosed spaces, the first thing that I label is the ceiling. The order of the annotations does not really matter, but one has to find what is more enjoyable or easier. Once I annotate the ceiling I label the walls and then all the other elements in the room, finishing with the floor. 


The procedure is simple: you need to take advantage of the corners of each object instead of clicking between them. Therefore, a window should have only 4 clicks. When the object is round or has curved edges it will need more points so that the appearance of the shape is not lost. When labeling a tree I use the mouse a lot. If the annotation of the tree is correct, one should be able to differentiate among a tree with leaves and one without. When I work on a picture I always think about the final result. I want the result to look beautiful when looking at the colored polygons. Before deciding that a picture is complete I hear an internal voice asking ``don't you see a chair over there? Doesn't that thing look like a pipe?'' As a result, before closing a picture, I always label a few more things.


In a picture, I always annotate in a row all the objects that are of the same type. This is a simple way to avoid making orthographic mistakes. 

I never label the objects that are reflected in a mirror, but it is interesting to notice how the reflection helps to identify some objects that are not clearly identifiable. Using some reasoning seems to have an important role when recognizing objects.


For a picture such as the following one it is important to be fully awake and, armed with all the energy that helps one start a new day. 
\begin{figure}[H]
\centering
\includegraphics[width=.9\linewidth]{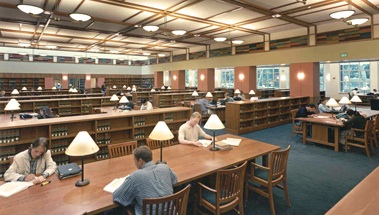}
\end{figure}

There are lots of elements close together, and some of them, specially on the picture background, are not clearly defined. It is very important to clearly differentiate between the tables and the shelves:

\begin{figure}[H]
\centering
\includegraphics[width=.9\linewidth]{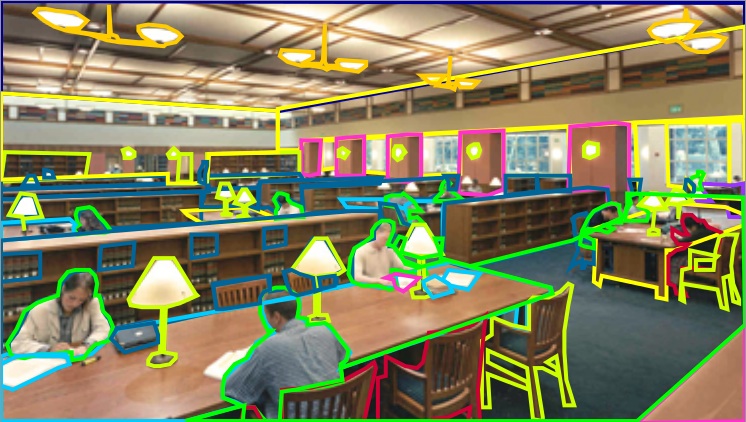}
\end{figure}



\subsection{Holes and wiry objects}

There are things that I like labeling: such as fences or handrails. I start by annotating the picture without worrying about them and I add them at the end. In the next example I labelled a handrail:
\begin{figure}[H]
\centering
\includegraphics[width=0.9\linewidth]{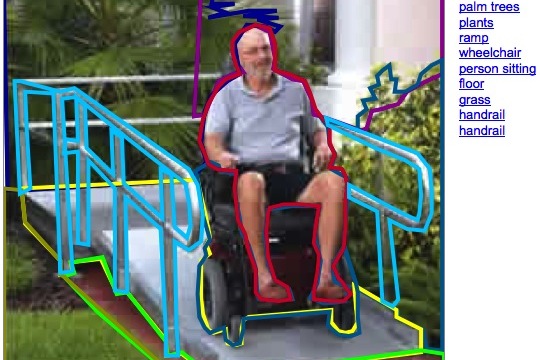}
\end{figure}
 I annotated the handrail by following all the bars so that one can see through the holes, providing a clear image of the object: 
 \vspace{-0.1in}
\begin{figure}[H]
\centering
\includegraphics[width=0.55\linewidth]{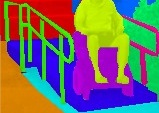}
\end{figure}

Sometimes, when labeling these types of objects, I need to delete the polygon once I finish it because the object mask contains exactly the opposite of what I wanted (i.e., the holes are marked as being the object). In those cases I need to think about doing exactly the contrary to what I wanted. The problem arises because I wanted to label the object without going over the same place twice, though that might be impossible when there are a lot of bars. 
Going over the same places multiple times may be easier even when there might be a way of labeling the object without repetitions. 

Labeling chairs can also be fun, because the legs have lots of bars that cross each other. It might not be of scientific interest but it looks beautiful and I do not just work, I also have fun. One scientist once told me: ``if it looks beautiful it means that it is correct.''


When a door is open I label what can be seen through it. In this way I collect a few more objects and it gives a sense of depth that might be interesting to have for some undefined goal. In a building, I label all the elements that are well differentiated and not just doors and windows. For instance, in some cases the columns are very distinct.

\begin{figure}[H]
\centering
\includegraphics[width=0.9\linewidth]{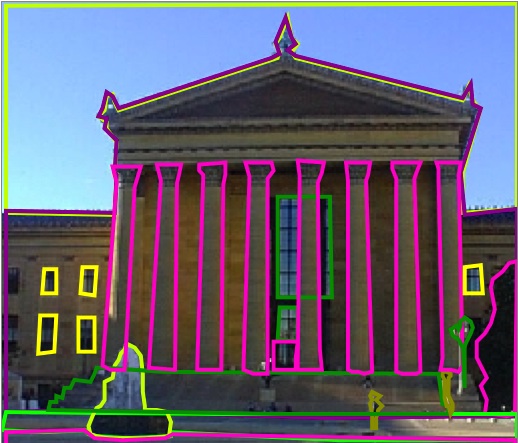}
\end{figure}


There are images that are very complicated and produce a great satisfaction when they are finished. For instance, in this image, the ring is the most difficult part. You want to avoid it looking like a solid block which would create a mask very difficult to identify and anti-aesthetic for the rest of the picture. There are also objects that I can not manage to recognize or whose names I don't know. In such a case it is better to pretend not to have seen them and skip to the next picture. 

\begin{figure}[H]
\centering
\includegraphics[width=1\linewidth]{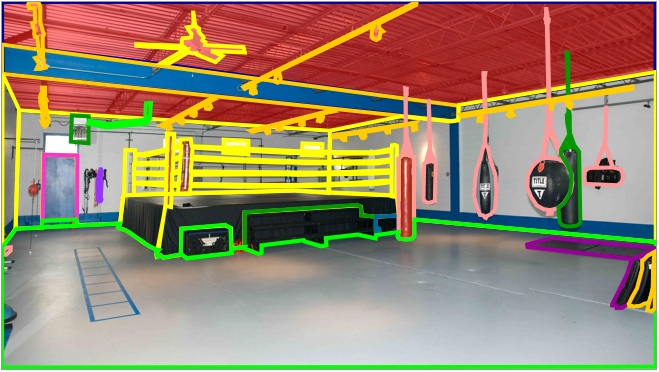}
\end{figure}


\subsection{Difficult images}

There are some pictures that are impossible for me. In order to understand them I would have to study engineering which could take a bit too long for me. The next picture is not the hardest picture but it is a good example. So we better skip this one ...
\begin{figure}[H]
\centering
\includegraphics[width=1\linewidth]{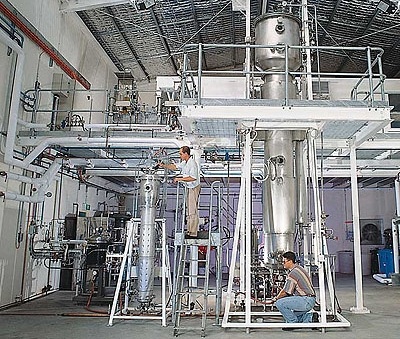}
\end{figure}


\newpage
If an object is occluded it is because another is occluding it. I always label both objects making sure that the edges coincide. An object starts where another ends. 


\begin{figure}[H]
\centering
\includegraphics[width=.9\linewidth]{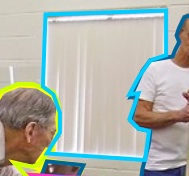}
\end{figure}


There are pictures that when I see I have to shout ``ooof'' and it is because I really have no idea how to start with them. But then I think that I have to do this picture for a special work and I just have to do it. After that reflection my mind becomes focused and it is like doing a job and not just a caprice.  Once I start with the first mouse click everything becomes easy and it is smooth sailing from there. 
 
 
Labeling requires zooming in and out. Zooming in helps to label small details but it can also create confusion. For instance, when I look at a magnified portion of a picture:
\begin{figure}[H]
\vspace{-0.1in}
\centering
\includegraphics[width=.07\textwidth]{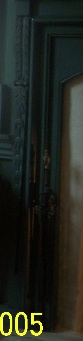}
\vspace{-0.1in}
\end{figure} while at first it seemed to be a door, I realized that it was part of a chimney once it returned to the original size. 
One can even see the tools typical of a chimney: 
\begin{figure}[H]
\centering
\includegraphics[width=0.9\linewidth]{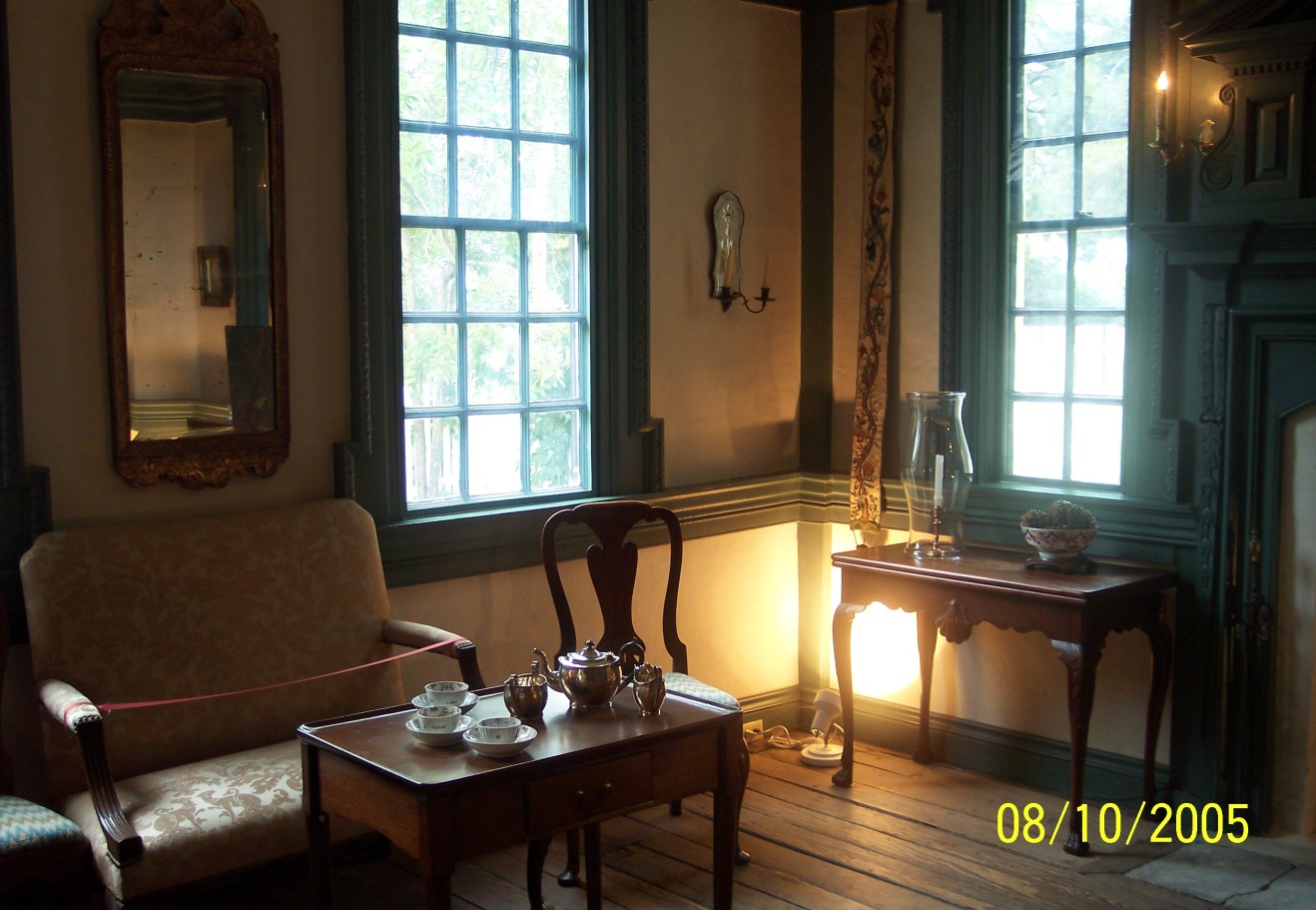}
\end{figure}
For this reason, before deciding that a picture is concluded, I always return to the original size to make sure that everything is correctly labeled. 
 
 


Here we have a picture that might seem complex at first sight. But it actually has only a few distinct elements and there is no reason for it to be difficult to label. I start by labeling the ground and the plants on top, then I label the ramps, and finally the water which is the easiest part:
\begin{figure}[H]
\centering
\includegraphics[width=0.48\linewidth]{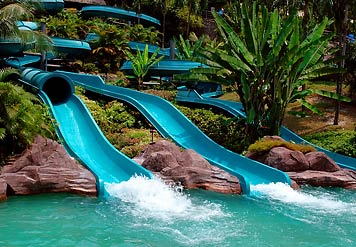}
\includegraphics[width=0.48\linewidth]{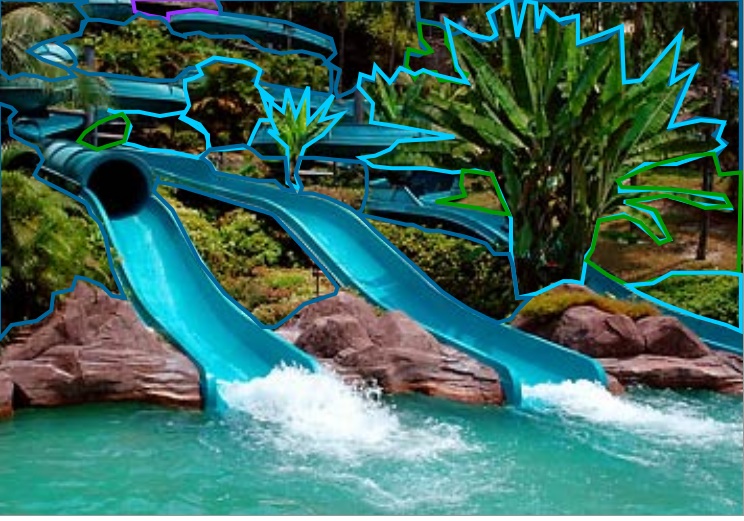}
\includegraphics[width=0.48\linewidth]{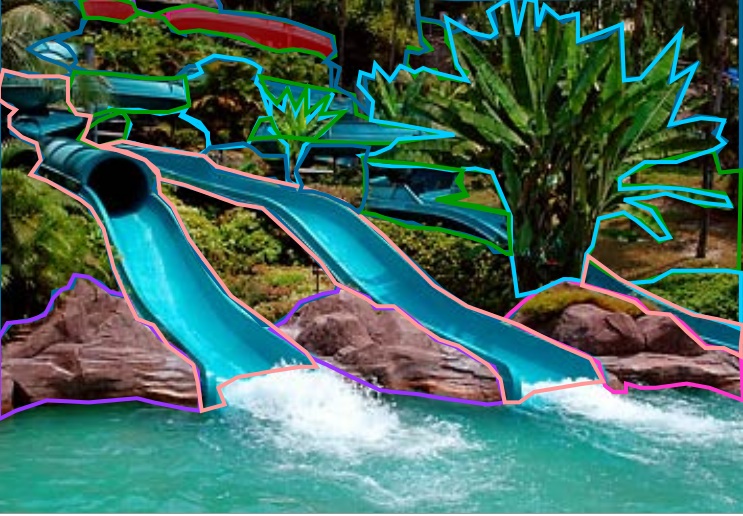}
\includegraphics[width=0.48\linewidth]{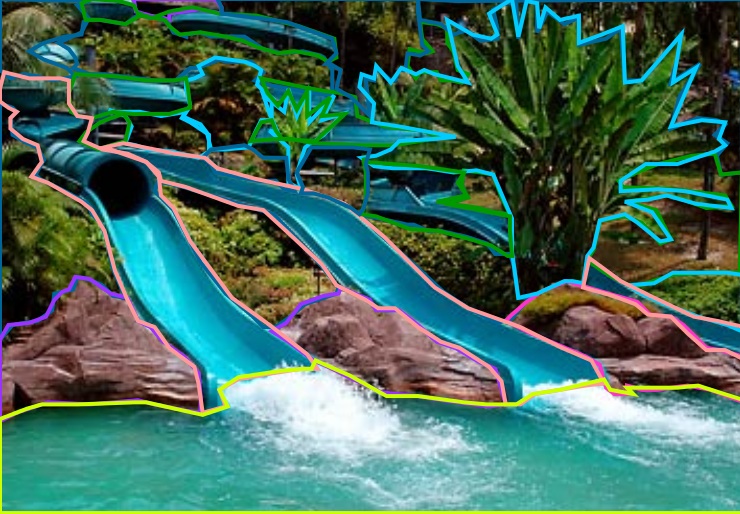}
\end{figure}

This labeling job has made me very observant.  I have found pictures that made me think ``if I had taken such a picture, then I would know what is everything.''  For instance, in a picture of a landscape, sometimes I do not know if I am seeing small trees or large bushes. The road to my home (I live far away from the city) passes by a landscape with lots of trees, wineries and even a small river. Now I look at this landscape in a different light because I want to recognize every tree, every bush and I try to think how each of those elements would look inside a photograph. Now I do not just look at things, I am also interested in knowing their names because in this job it is not good enough to give a description of what they are useful for.



This next picture has lots of elements, but as they are well defined and easy to identify, labeling this picture does not represent a challenge although it requires a laborious job.
\begin{figure}[H]
\centering
\includegraphics[width=.9\linewidth]{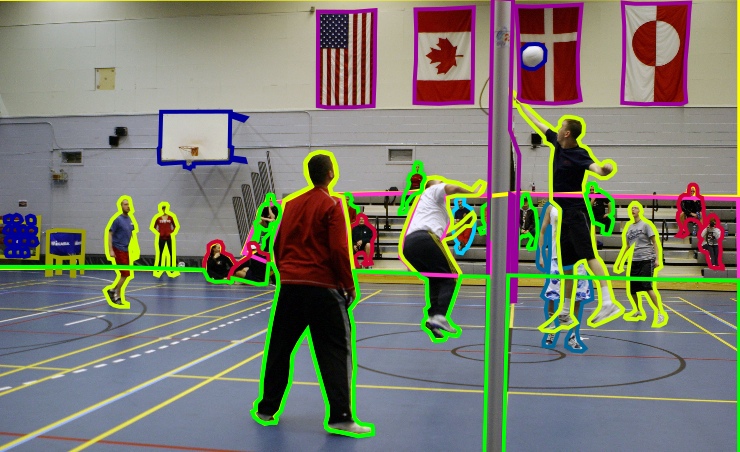}
\end{figure}

I find the next picture more complex than the previous one. There are lots of elements piled up that are difficult to identify and to differentiate. For this reason, one has to start labeling the large surfaces such as the ceiling and the floor, and then add, little by little, all the other elements that are recognizable. The most important issue that I encounter that makes me catalogue an image as being difficult is that I do not know the name of several of the objects. 


\begin{figure}[H]
\centering
\includegraphics[width=.9\linewidth]{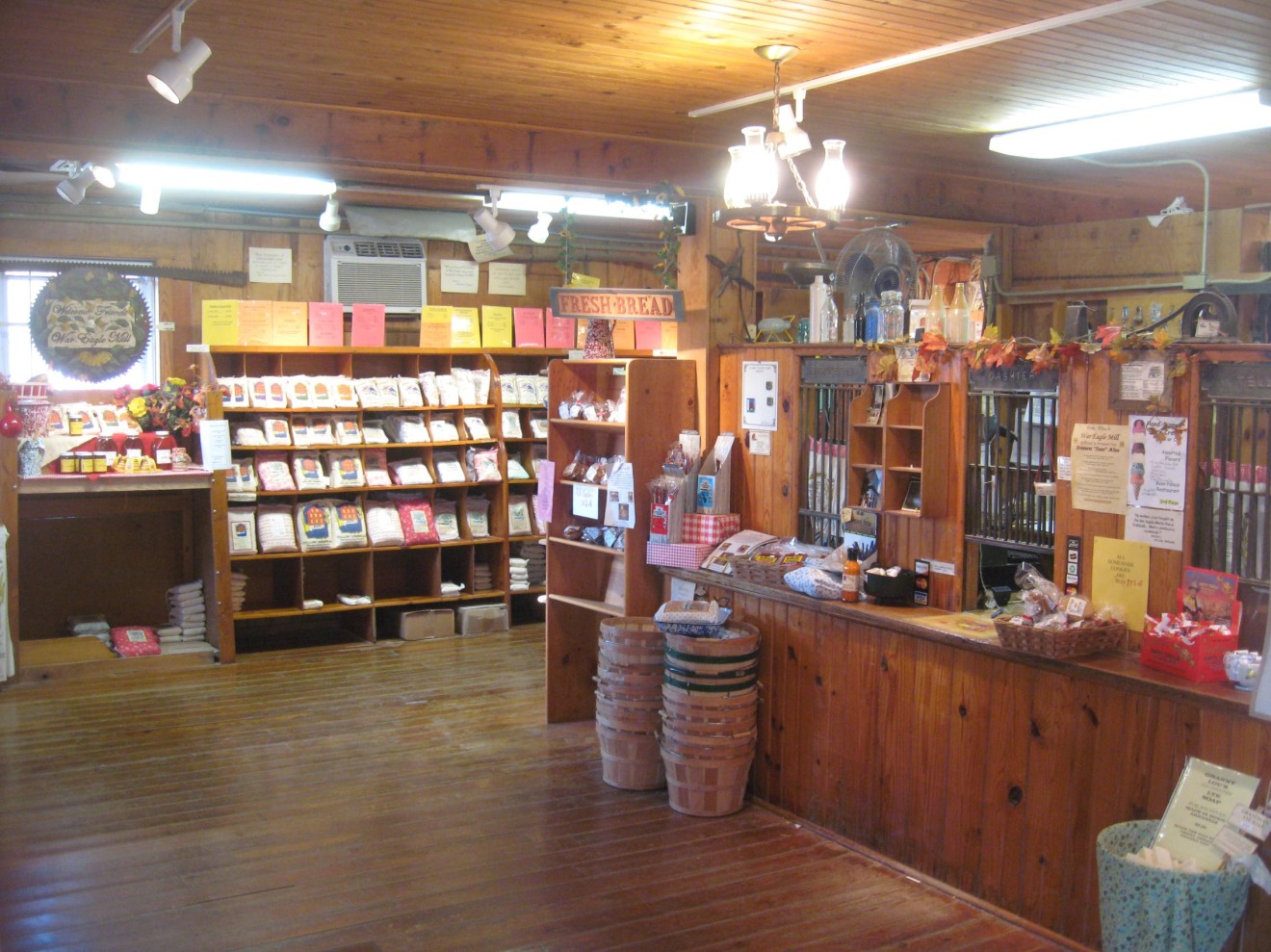}
\end{figure}

I always annotate walls with different orientations as being different instances even if they are connected. For instance, in a picture of a room one is quite likely to see three walls and each one will be annotated separately. Similar to a set of objects in which each instance would be labeled as a single item, the walls can also be separated. For walls the separation corresponds to different planes. I think it is interesting to annotate different wall orientations as different instances.



\vspace{-.1in}
\begin{figure}[H]
\centering
\includegraphics[width=.9\linewidth]{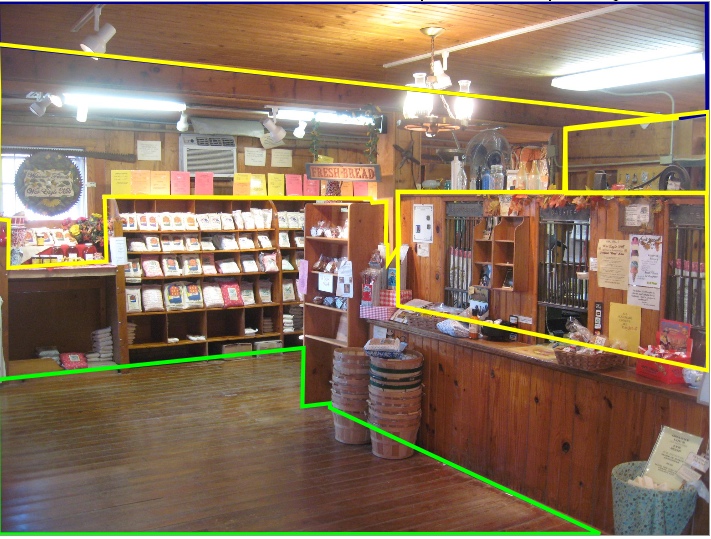}
\end{figure}
Then, I like labeling the furniture and other big elements, ending with the smallest ones that normally are supported by the big ones. 

\vspace{-.1in}
\begin{figure}[H]
\centering
\includegraphics[width=.9\linewidth]{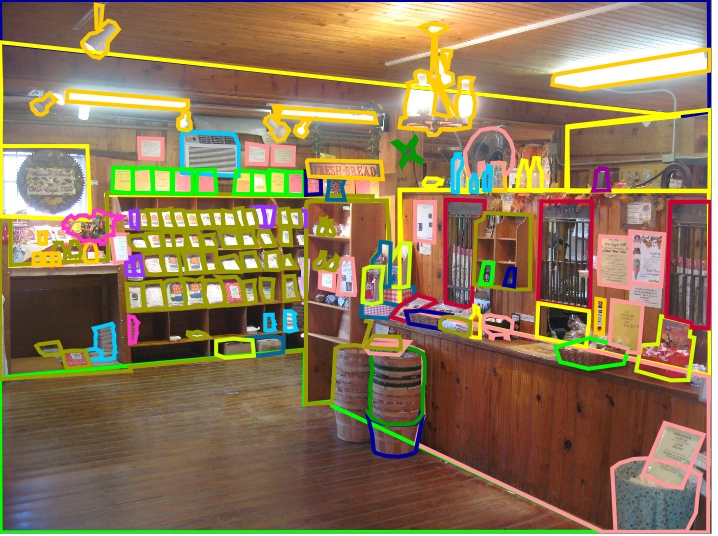}
\end{figure}

It is good to try to find the largest number of elements and avoid using plurals. In this image we can see that there are very few different types of objects, but it is a picture with lots of instances to annotate as we are trying to avoid using plurals.  




\begin{figure}[H]
\centering
\includegraphics[width=0.9\linewidth]{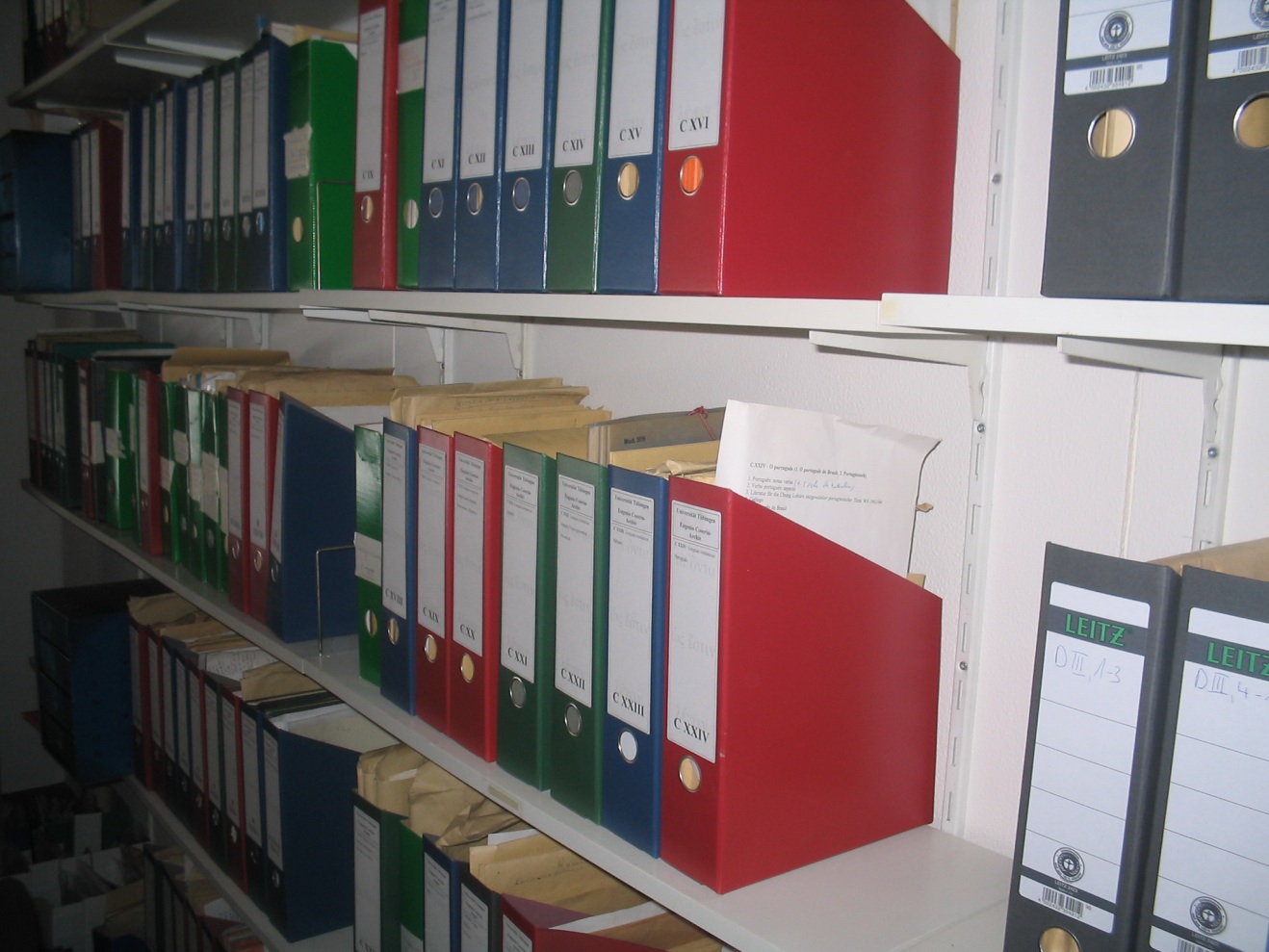}
\end{figure}
When an object is not completely visible because there is another one that hides it, then I write inside the object name the word ``occluded.'' However, I do not always add the word ``occluded'' to the object name. For instance, when I see books or folders (as in the next picture) that are in their natural place so that you only see the book spines, I never write the word ``occluded''. I do not use the word ``occluded'' when these objects are occluded because that is their natural way of appearing.
\begin{figure}[H]
\includegraphics[width=0.89\linewidth]{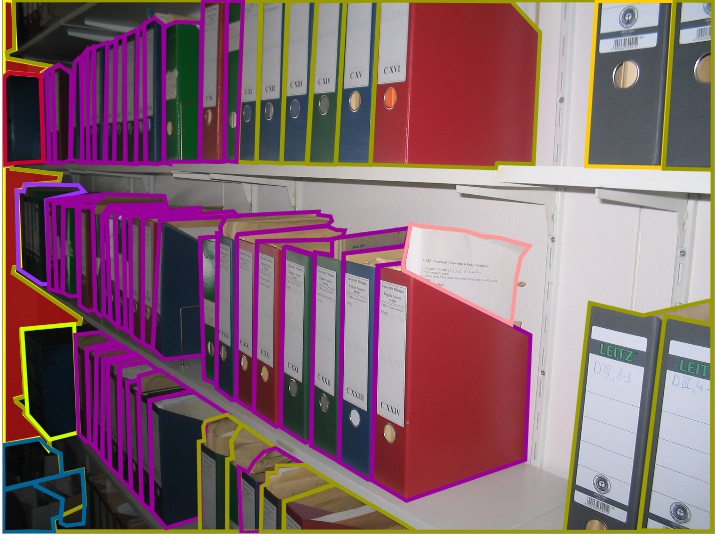}
\includegraphics[width=0.08\linewidth]{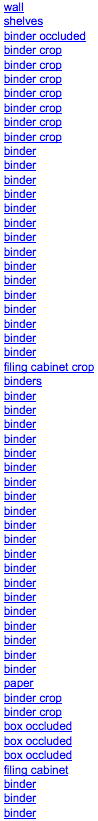}
\end{figure}

During the time that I have been doing image annotations, I have encountered several interesting cases that have made me think, rectify, and deduce what I was seeing. But once I decided to write this little article, many of those anecdotes have disappeared from my mind and I have only been able to explain the situations that I was encountering since I started writing. But I do not discard that I might continue adding new experiences that I will continue collecting day after day.



When I was proposed to work with LabelMe, I found the task interesting because it was something that I'd never done before. The beginning was easy because the pictures that I was given to label were very simple. They contained very specific things that were easy to recognize. But, little by little, the pictures they sent me became more and more complex and suddenly nothing seemed easy. But when you devote several hours a day to a job, you start mastering it and the difficulty has to be very large in order to become impossible to do. However, even after all my labeling experience, I still find images that I do not know how to annotate.


\newpage

The next picture represents a big challenge. What is on the right side?

\begin{figure}[H]
\centering
\includegraphics[width=0.9\linewidth]{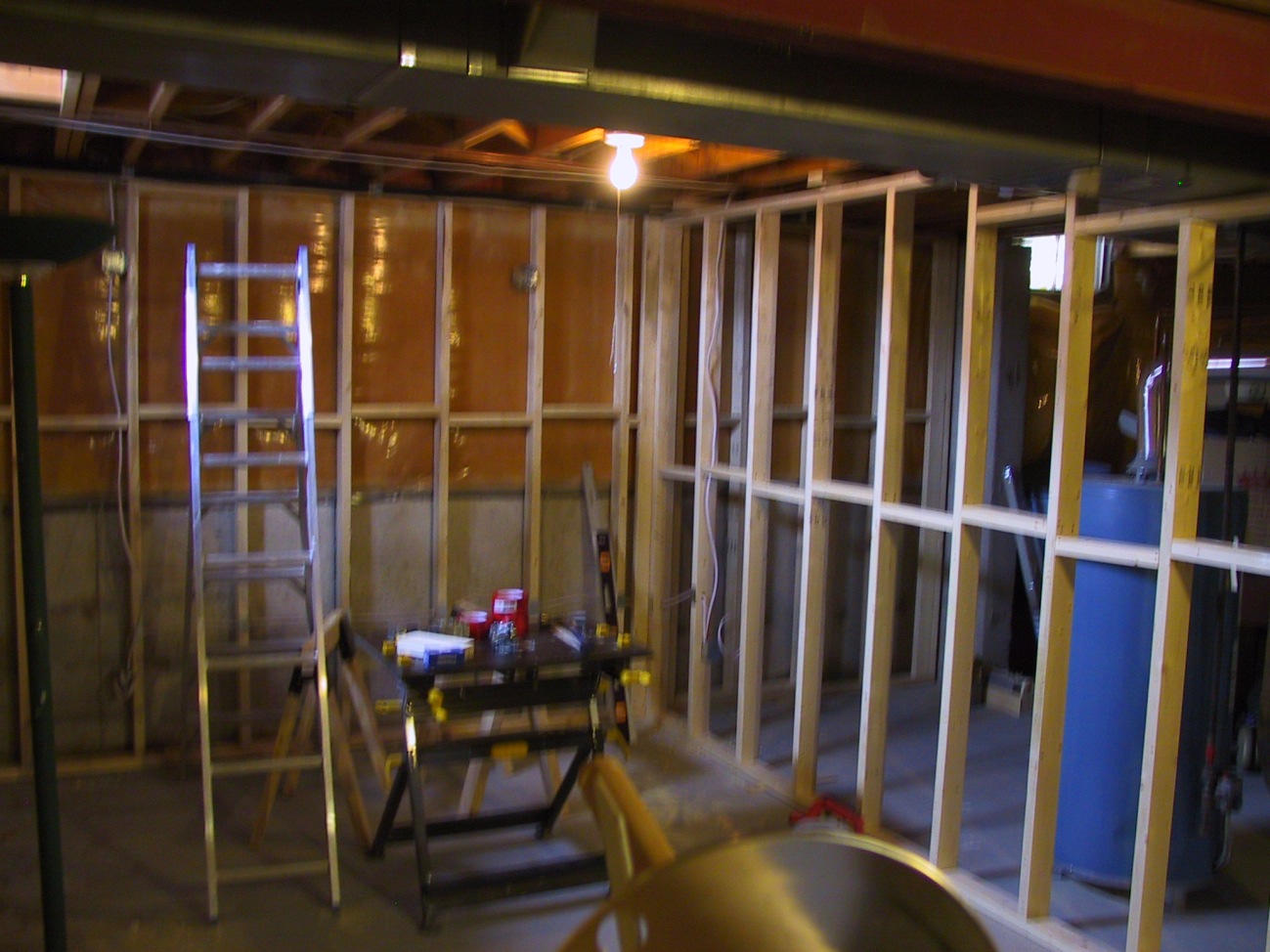}
\end{figure}
I can see the ceiling, a wall and a ladder, but I do not know how to annotate what is on the right side of the picture. Maybe I just need to admit that I can not solve this picture in an easy and fast way. But if I was forced to label it then I would proceed as follows: I would start ignoring the unfinished wall that will split the room and I would extend the walls and ceiling. Then, at the end, I would label the wood of the splitting wall in such a way that the object mask will allow seeing what is behind, just as in the picture. I have no idea about what is the object that is in the frontal plane of this picture.

And this picture...
\begin{figure}[H]
\centering
\includegraphics[width=0.9\linewidth]{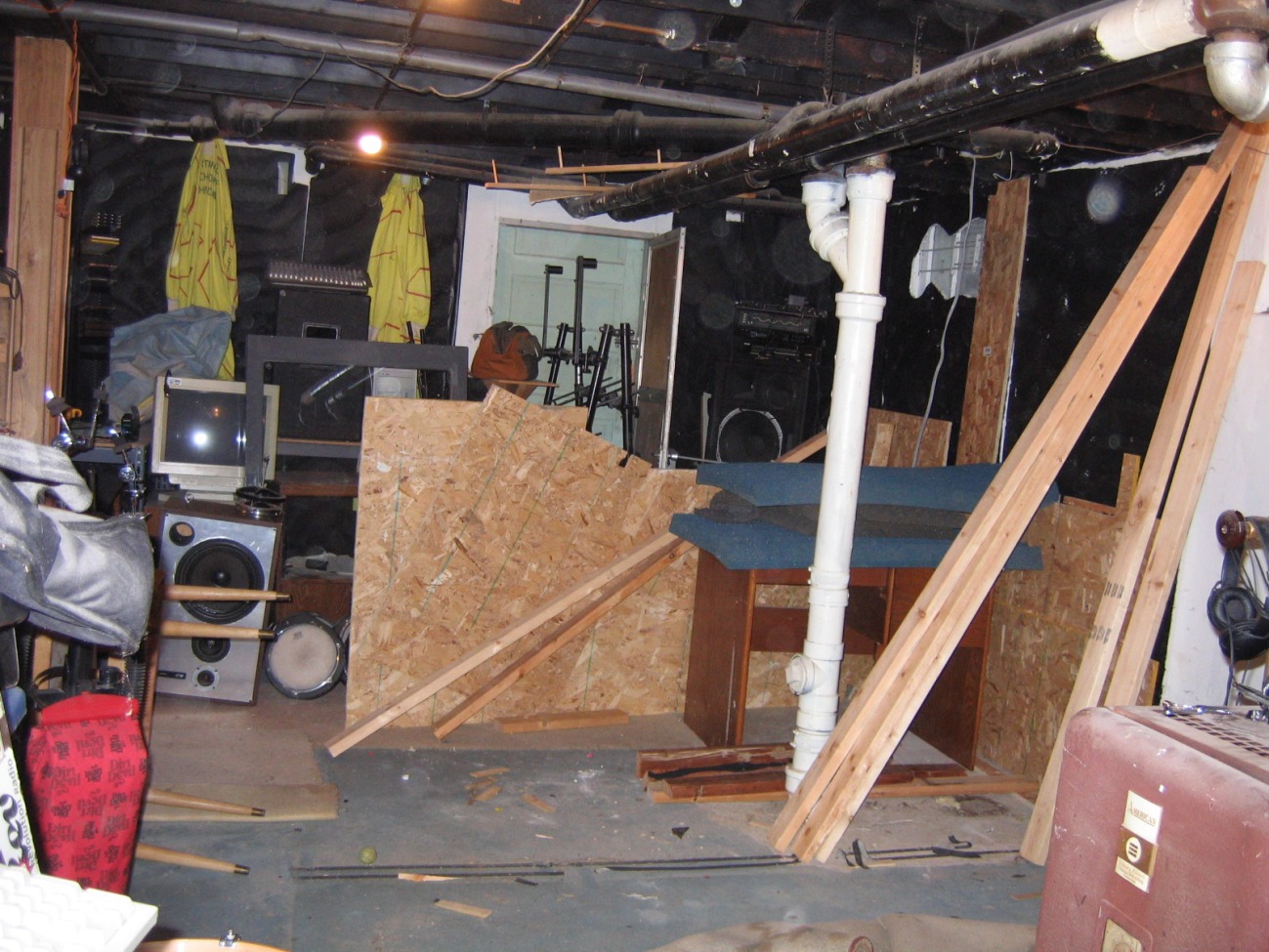}
\end{figure}
...it is such a mess that it seems  the mind does not want to make the fight to split every element. But, as it was the case with the previous picture, it would be possible to annotate the image if you found yourself with the duty to do it. The true problem appears when one does not recognize what an object is.


In this other picture, I do not know the name for most of the things in the scene. It looks like a lab, but how would you name the tables? Would you call them ``lab tables'', or ``work benches''? I generally skip any picture that I find difficult to label right from the beginning when I open it. I know this is a bad thing to do, but there are so many pictures and so much work...
\begin{figure}[H]
\centering
\includegraphics[width=1\linewidth]{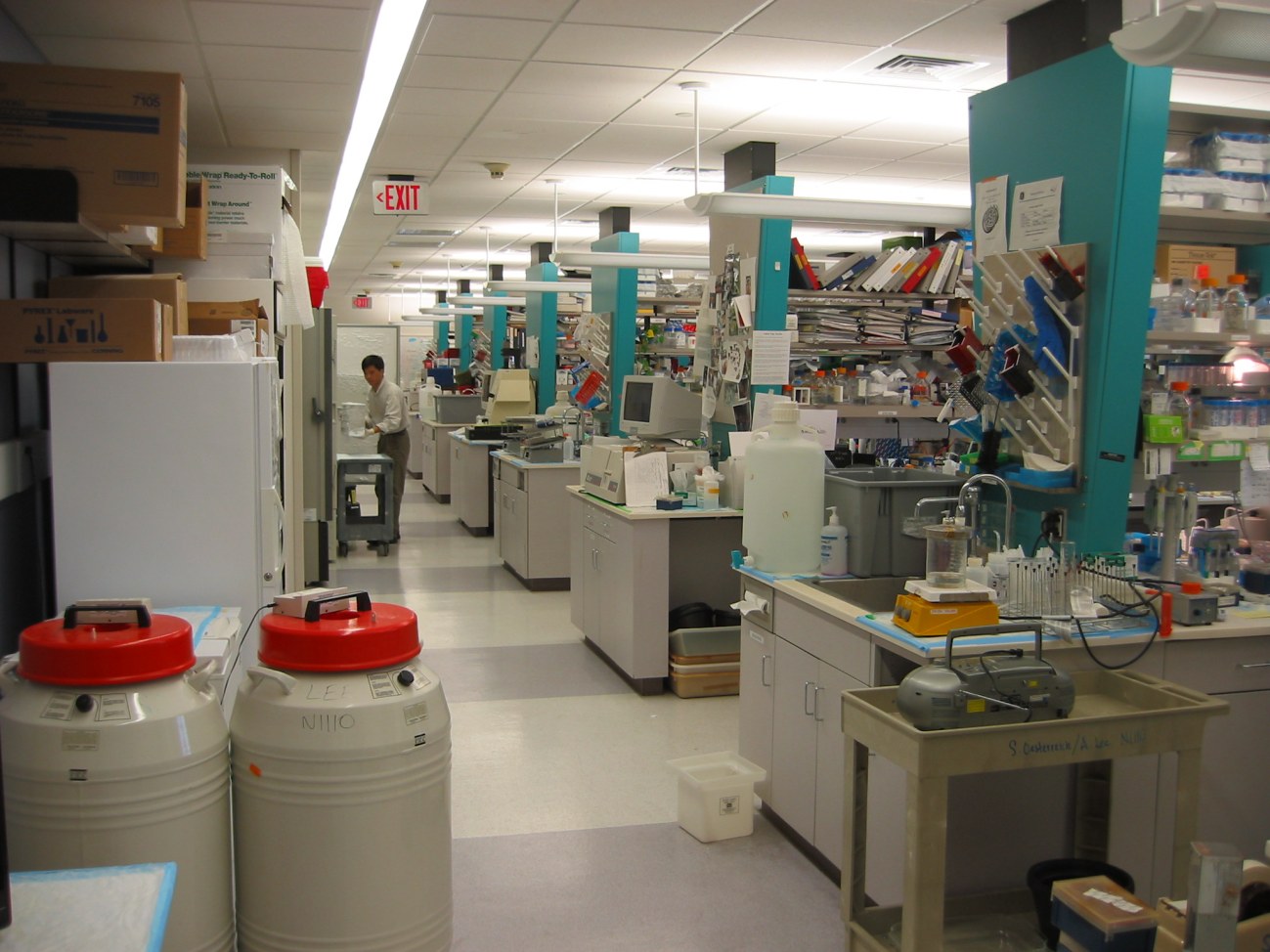}
\end{figure}

If we continue with pictures of labs, there are other examples that are even worse:
\begin{figure}[H]
\centering
\includegraphics[width=1\linewidth]{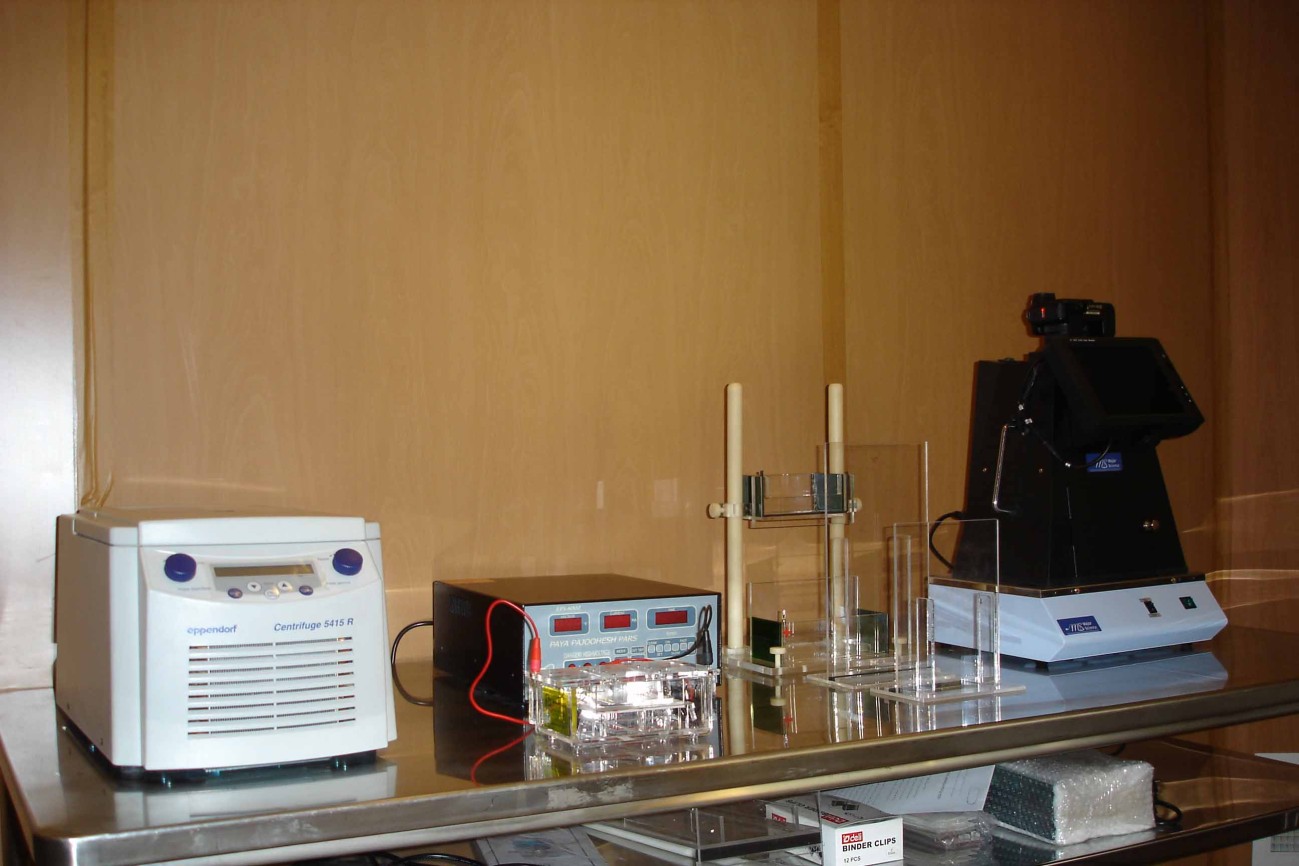}
\end{figure}
It might seem easy, but if I do not know what these objects are, how can I label them?

In the next picture, I do not even know what it is that I am seeing. These pictures make me nervous because I do not like feeling that I can not solve a problem. I look at this picture, but I am not able to understand it.
\begin{figure}[H]
\centering
\includegraphics[width=1\linewidth]{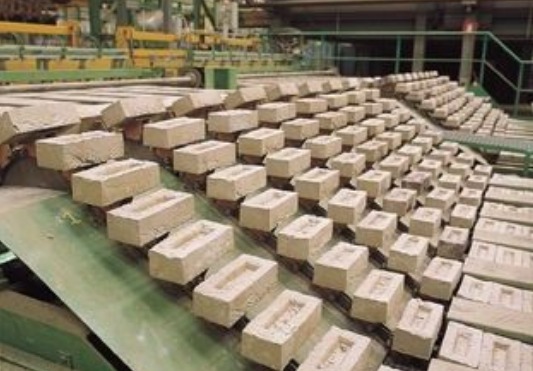}
\end{figure}
After looking at several other images within the same set of pictures I have arrived at the conclusion that the blocks are bricks. Sometimes, to understand a picture, one has to look at pictures within the same set as it can help to clarify many things. What you can not recognize in one picture might be easy to see in another picture. 

\subsection{Just nearly impossibly difficult}

There are some images that are just plain difficult. What should I do with the next picture? 
\begin{figure}[H]
\centering
\includegraphics[width=1\linewidth]{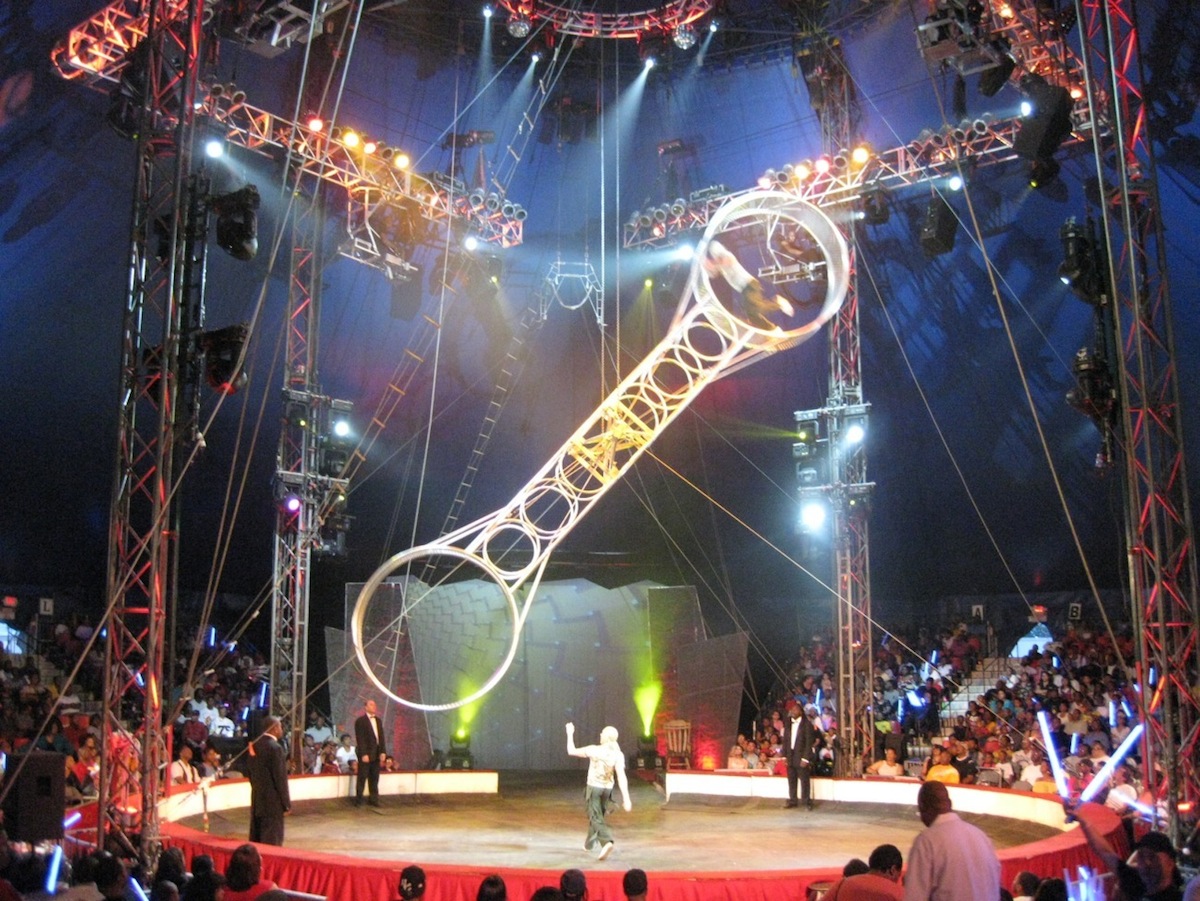}
\end{figure}
I am not sure what I am seeing here and what I can see, I do not know what to call it. I do not know what to call the background either. Is that a wall? a cloth? There are many elements that are well defined but I do not know how to catalogue them.

The pictures where there are a lot of people as the main subject of the photograph are very complex. As I do not want to use plurals, they become almost impossible to label as they are not clearly defined. It is hard to see where one person ends and another starts. And when the picture is inside a club with strange lighting, then it gets even more complicated as is the case with the next picture:
\begin{figure}[H]
\centering
\includegraphics[width=0.9\linewidth]{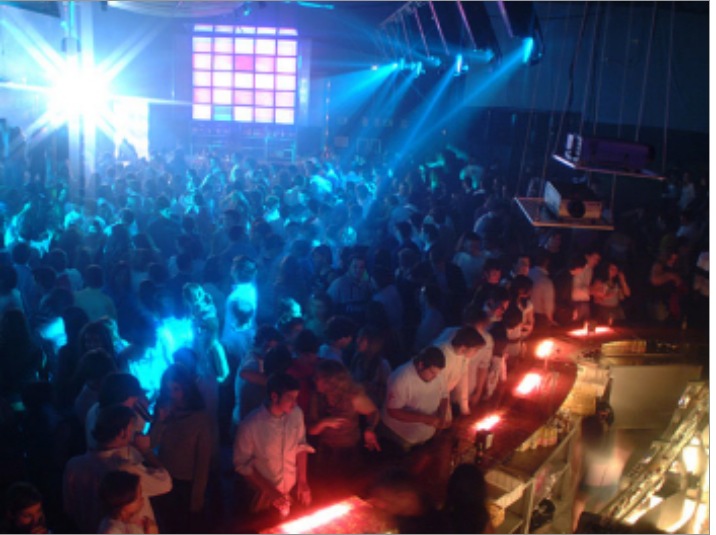}
\end{figure}

In this next picture it is very easy to see the people, but the challenge is in the background. I can see the lights. Therefore I assume the lights are on the ceiling. But I do not know where the ceiling starts and ends. I do not know how to annotate that.
\begin{figure}[H]
\centering
\includegraphics[width=0.9\linewidth]{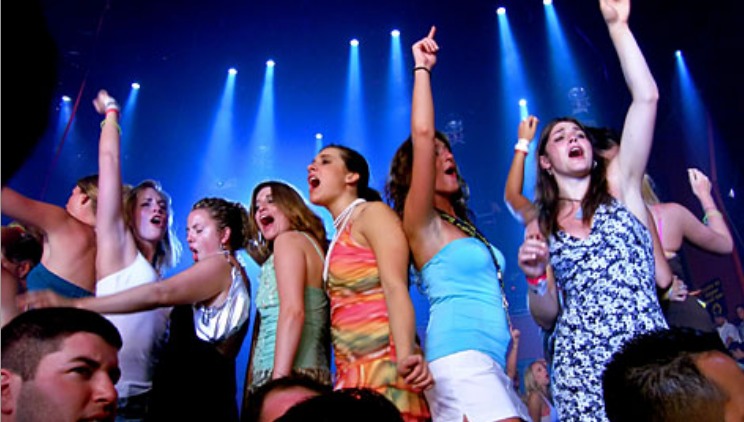}
\end{figure}

And this picture is spectacular, what is on those shelves?

\begin{figure}[H]
\centering
\includegraphics[width=0.9\linewidth]{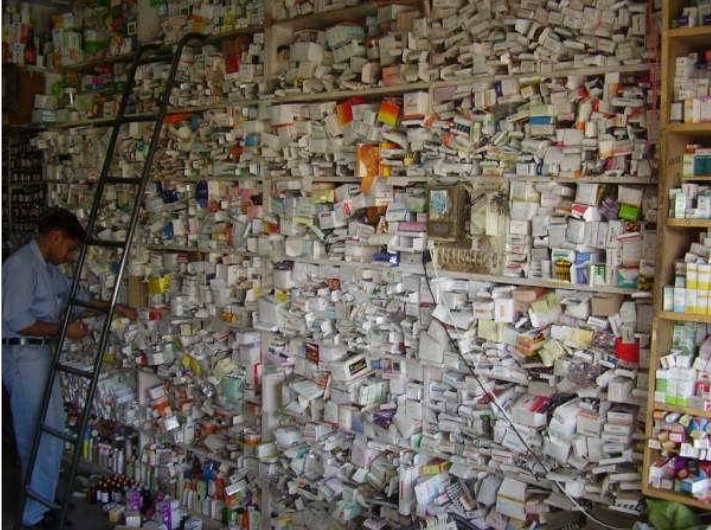}
\end{figure}


Other pictures are easy to understand visually, but I cannot do them because I do not know what the objects inside are. What are the objects in the next picture? They are not even machines.
\begin{figure}[H]
\centering
\includegraphics[width=1\linewidth]{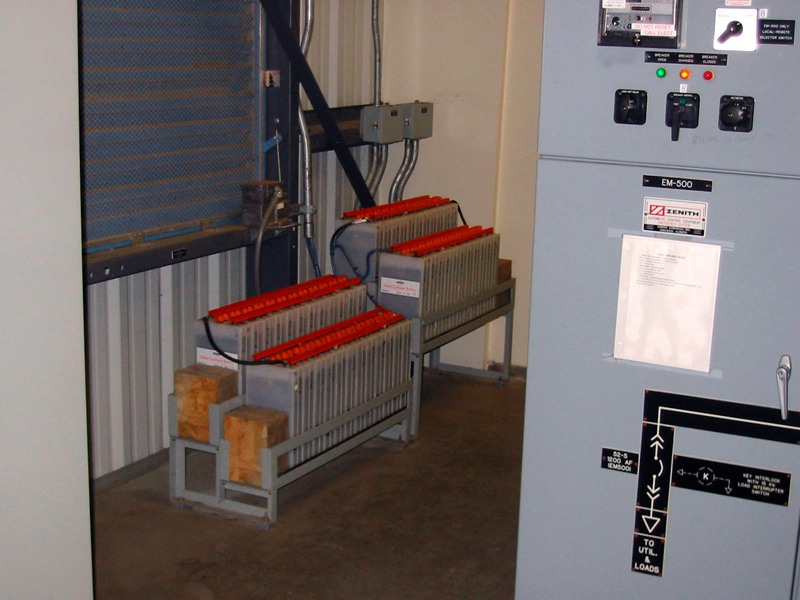}
\end{figure}

In a picture, the hardest part to annotate is what is far in the background because it is not very well defined and it is hard to see all the elements:
\begin{figure}[H]
\centering
\includegraphics[width=1\linewidth]{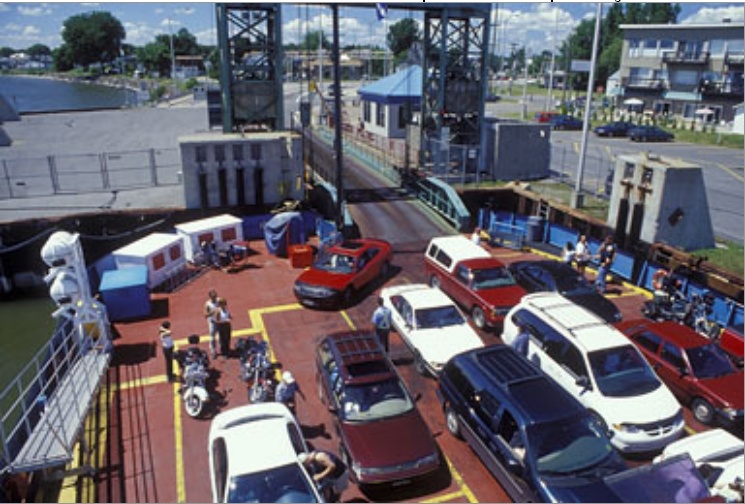}
\end{figure}

I always try to simplify as much as I can, but I do not know if it is very correct.


Generally, I have a tendency to skip pictures with lots of elements including ones whose names I don't know. But sometimes I work on those hard pictures as it is a way of making other pictures feel easier later.


\newpage

The next picture is a very good example of a picture that I have liked to skip:
\begin{figure}[H]
\centering
\includegraphics[width=1\linewidth]{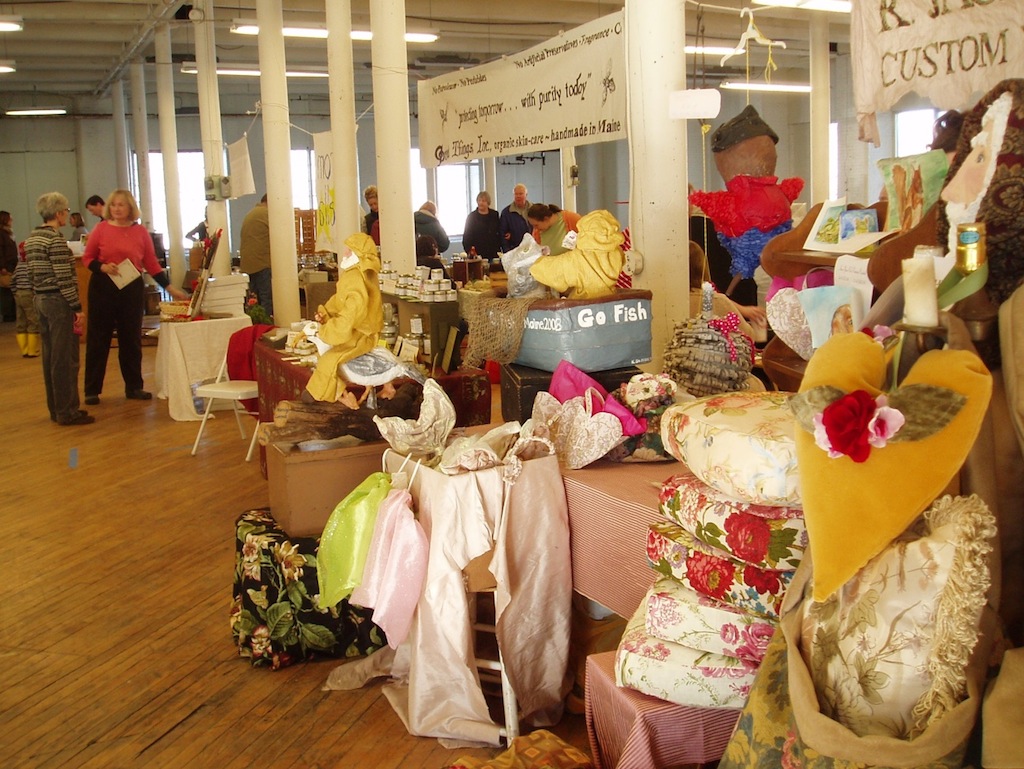}
\end{figure}

But I decided to do it. As I move forward labeling this image, I realized that the best way of proceeding with it was to label first the floor and walls, making sure that the walls touch the floor. As the boundaries of the walls and floor are not visible, one has to take into account the perspective of the scene. By labeling first the walls and floor I can label everything else without being worried about leaving any portion of the image without a label. Once I finished with this image there were more than 100 labels. 
\begin{figure}[H]
\centering
\includegraphics[width=1\linewidth]{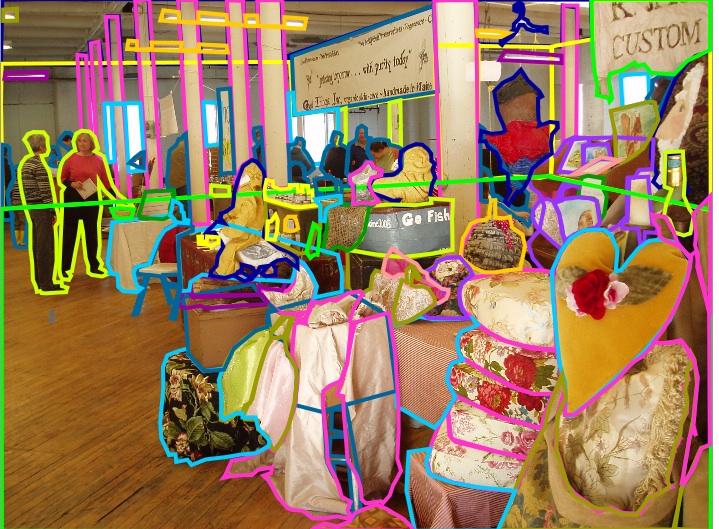}
\end{figure}

In general, pictures taken inside markets are among the hardest ones to annotate due to the large variability of object types that they contain. 
\begin{figure}[H]
\centering
\includegraphics[width=1\linewidth]{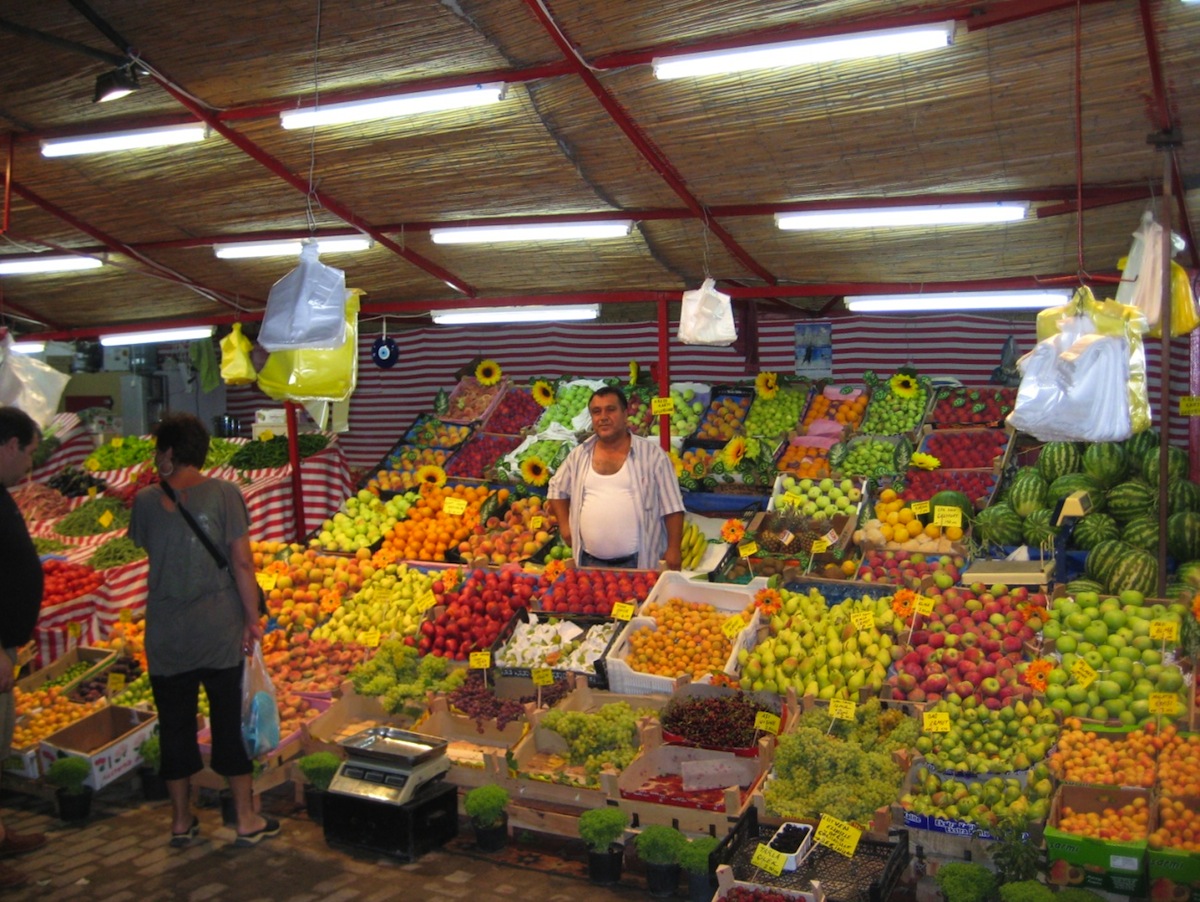}
\end{figure}

Probably, if I was not writing this paper I would have never annotated this picture, but, as I wanted to write about the fact that some images are impossible to label, I realized that labeling this image was not impossible. It just demands a lot of patience and the desire to do a good job.

\begin{figure}[H]
\centering
\includegraphics[width=1\linewidth]{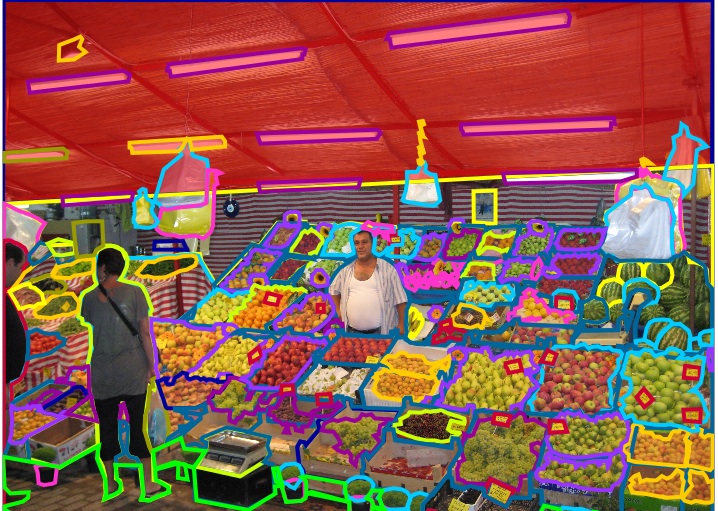}
\end{figure}
These types of pictures are better done at the start of a day, as I am not tired and I place more importance on quality rather than quantity. It is really fantastic to finish a picture like the previous one. Not all the images need to be simple and with few elements. However, it is almost impossible to do two images like the previous one in a row. 

\subsection{Depth planes}

In the next picture, the complexity of the image is due to the different depth planes made by the walls and arches. The walls are also partially visible through the arches. If we want to correctly represent this image giving to it all the depth it has, then it is essential to annotate what we see through the arches. 
\vspace{-.1in}
\begin{figure}[H]
\centering
\includegraphics[width=.9\linewidth]{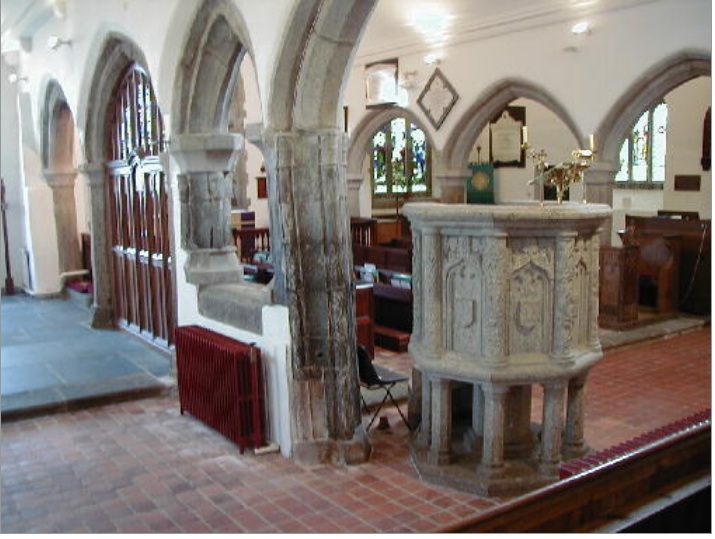}
\end{figure}
\vspace{-.1in}
In this case I am going step by step to illustrate how I annotate this image. First I start with the two main walls:
\vspace{-.1in}
\begin{figure}[H]
\centering
\includegraphics[width=0.9\linewidth]{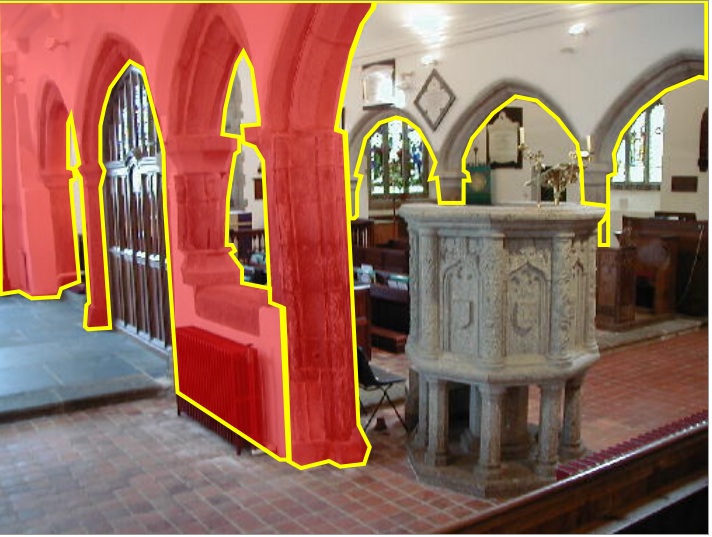}
\end{figure}
\vspace{-.1in}
Then I label the walls that I see through the arches and all the large objects that are easy to recognize. Many times I leave the floor for last, but in this case it is useful to study the floor.
\begin{figure}[H]
\centering
\includegraphics[width=0.9\linewidth]{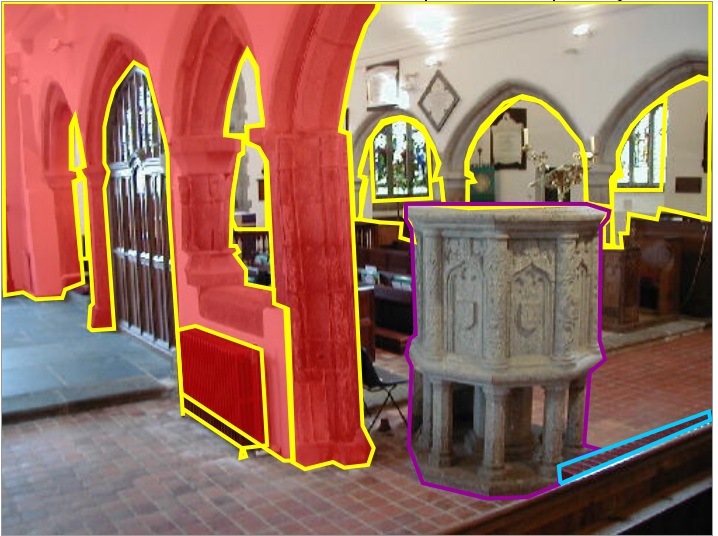}
\end{figure}
Here is the final annotated image. There is always the possibility that somebody might recognize something that I was not able to label.
\begin{figure}[H]
\centering
\includegraphics[width=0.9\linewidth]{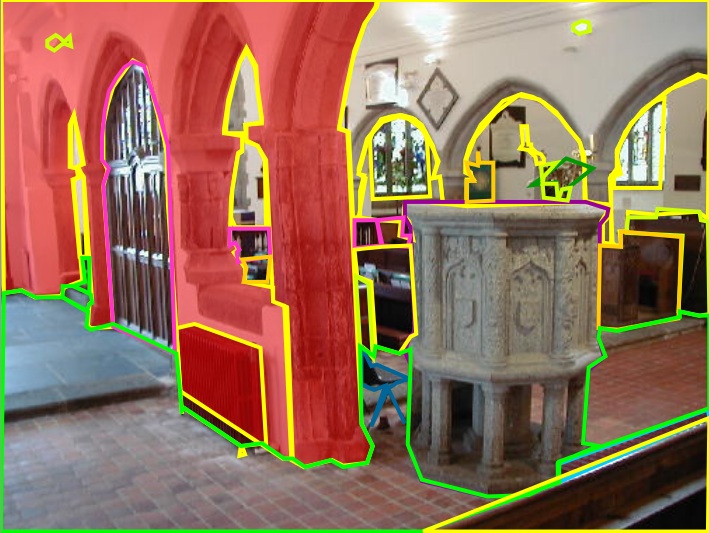}
\end{figure}
\vspace{-.1in}


\subsection{Kitchens}

Pictures taken inside kitchens are very hard. Here I refer to pictures of kitchen inside restaurants. Fortunately, kitchens inside homes have fewer elements and generally objects inside a home kitchen are well organized or stored inside cabinets, and they are usually well defined and easy to recognize. 

But look at these two pictures:

\vspace{-.1in}
\begin{figure}[H]
\centering
\includegraphics[width=0.49\linewidth]{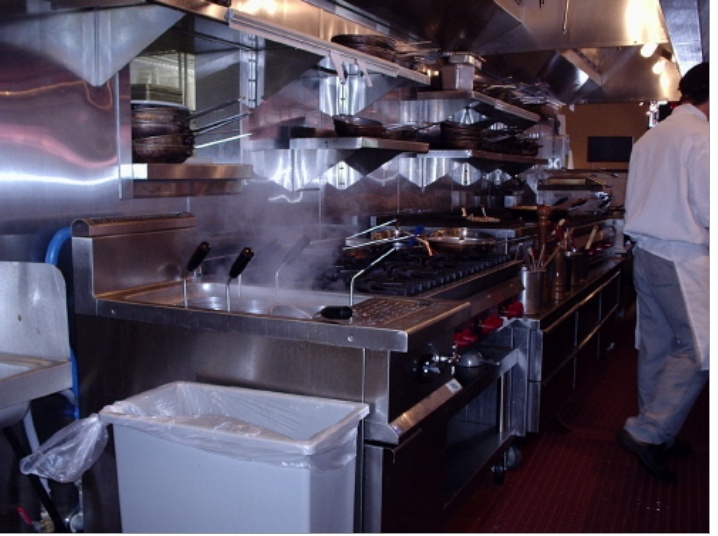}
\includegraphics[width=0.49\linewidth]{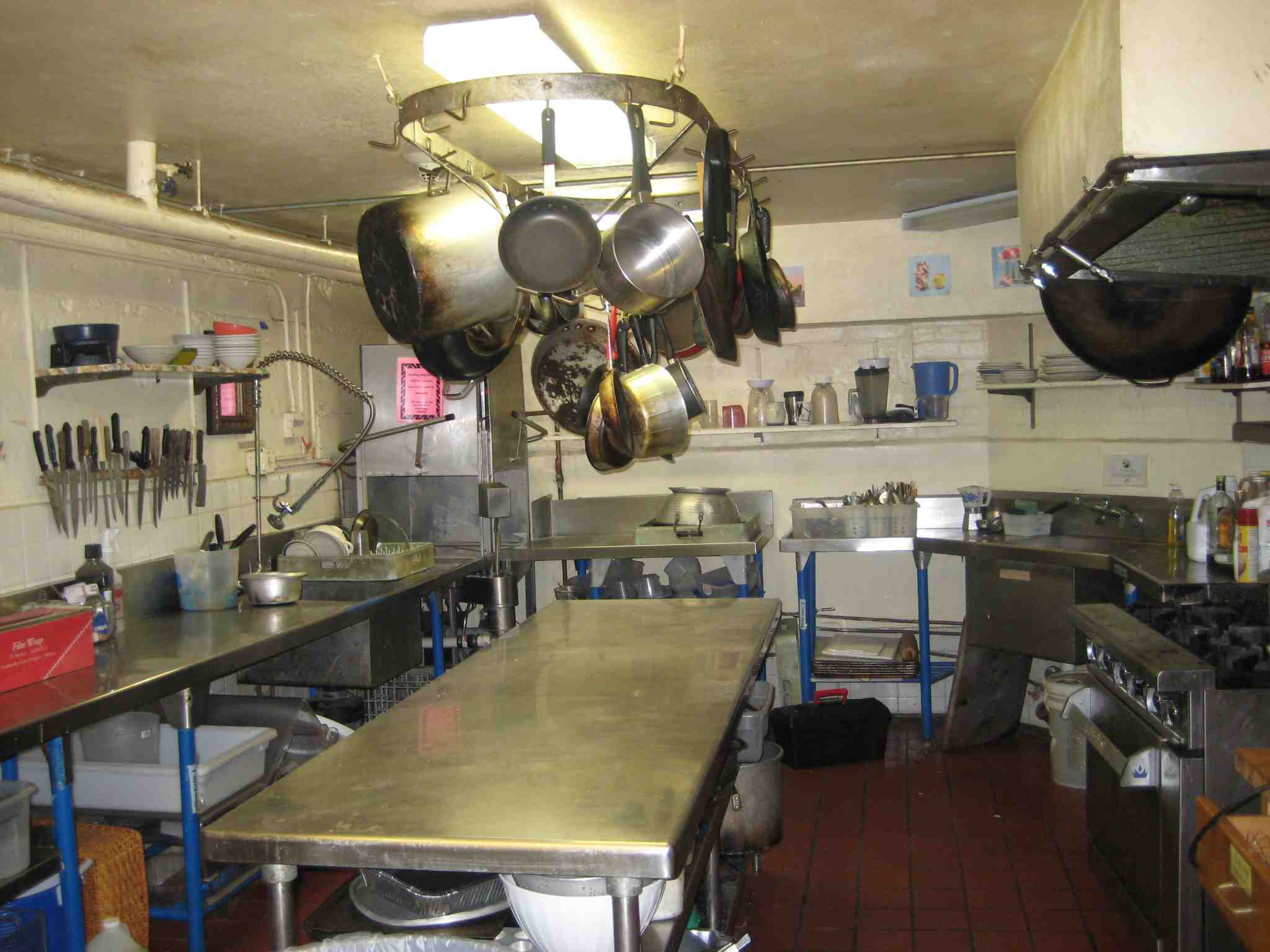}
\end{figure}
\vspace{-.1in}
Inside these pictures above, there are so many things, all visible and piled up. Pictures like those ones are very difficult and require a lot of patience.  

\subsection{Crystals, windows and mirrors}

In the next picture, all the elements are so clear and easy to identify that labeling this image is easy and very gratifying. In this picture I have only encountered two dilemmas. As you can see, on top of one of the tables we can see a few transparent containers. In one of them there is a cake, and in another one there seem to be some cookies. I did not know how to name those objects: should I call it a ``cake'' or should I call it ``container''?  I decided to name it a ``container'' because I do not label the things that are visible behind a crystal or something transparent. This is a rule that I always follow. The second issue that I have encountered is the chair that is hanging on the wall. If my goal is to teach the concept of what a chair is, then having the chair on a wall leads to difficulties of understanding, but that is what it is.

\begin{figure}[H]
\centering
\includegraphics[width=1\linewidth]{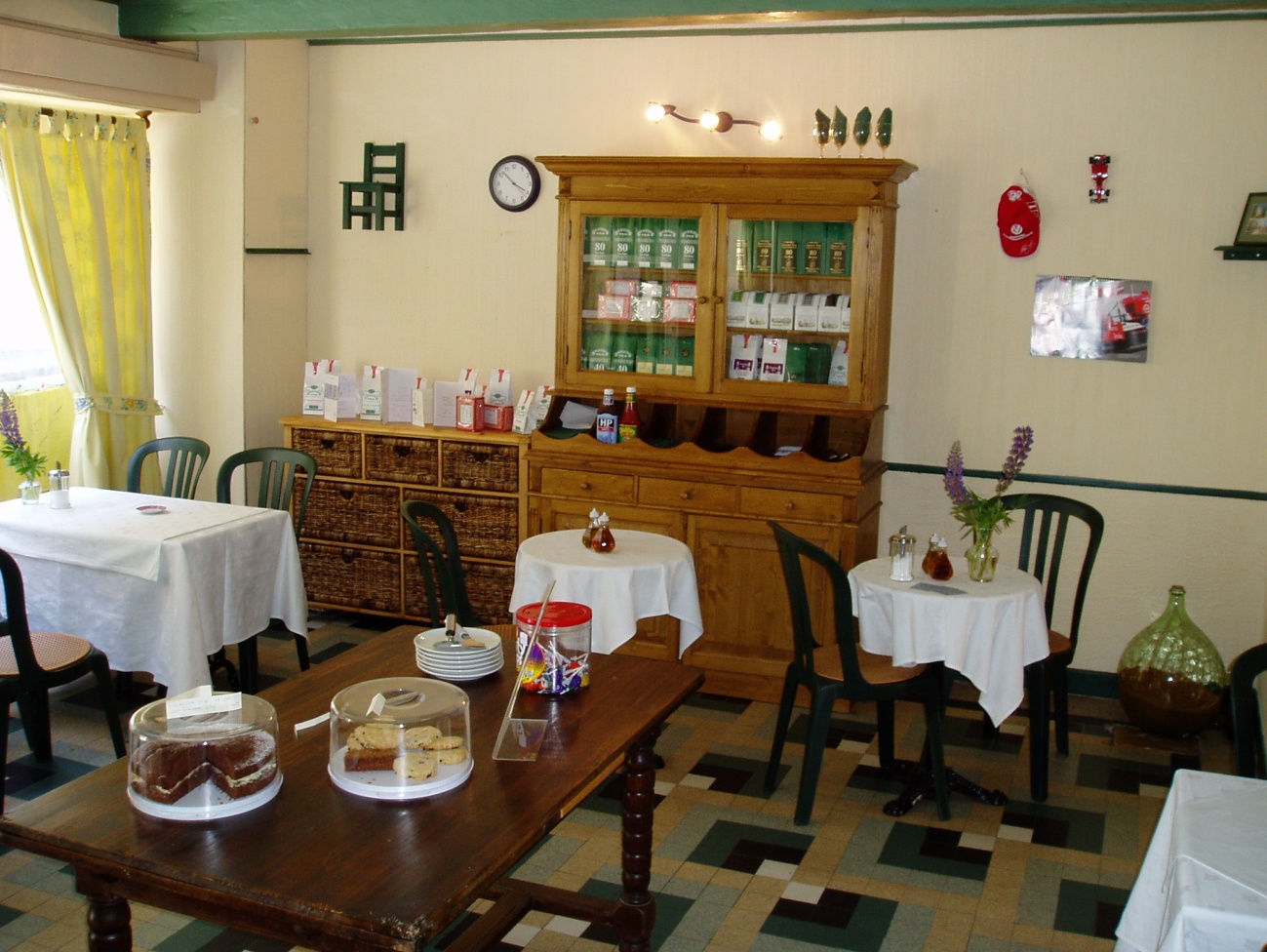}
\end{figure}
Once everything has been labeled the result is:
\begin{figure}[H]
\centering
\includegraphics[width=1\linewidth]{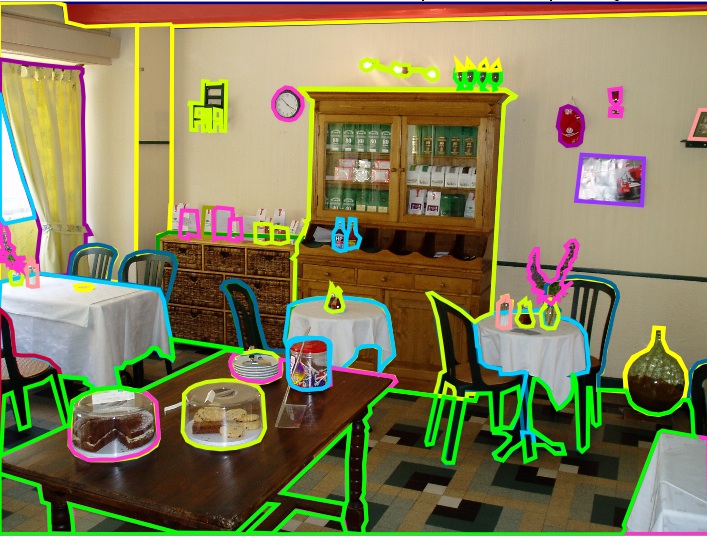}
\end{figure}

As you can see, inside the cabinet there are a number of bags and none of them are annotated. The reason is that I never label the objects that are inside cabinets and that are visible behind glass doors. I also never label anything behind a closed window. I am not sure if this is the right thing to do, but in many cases one has to adopt some criteria (unless somebody corrects me). There are so many open windows that I ask myself: why should I also label the objects that are behind closed windows?  

There are other places were I also use my own criteria. For instance, when a window is totally occluded by a curtain, when I label it I call it ``curtain'' and I do not use the word ``window.'' Another example of an arbitrary decision is when I label flowers. If the flowers are in a pot I call it ``plant.'' However, if they flowers are cut and inside a jar then I call them ``flowers'' because, in my opinion, once they are cut they are not plants anymore.



The next picture confused me:
\begin{figure}[H]
\centering
\includegraphics[width=1\linewidth]{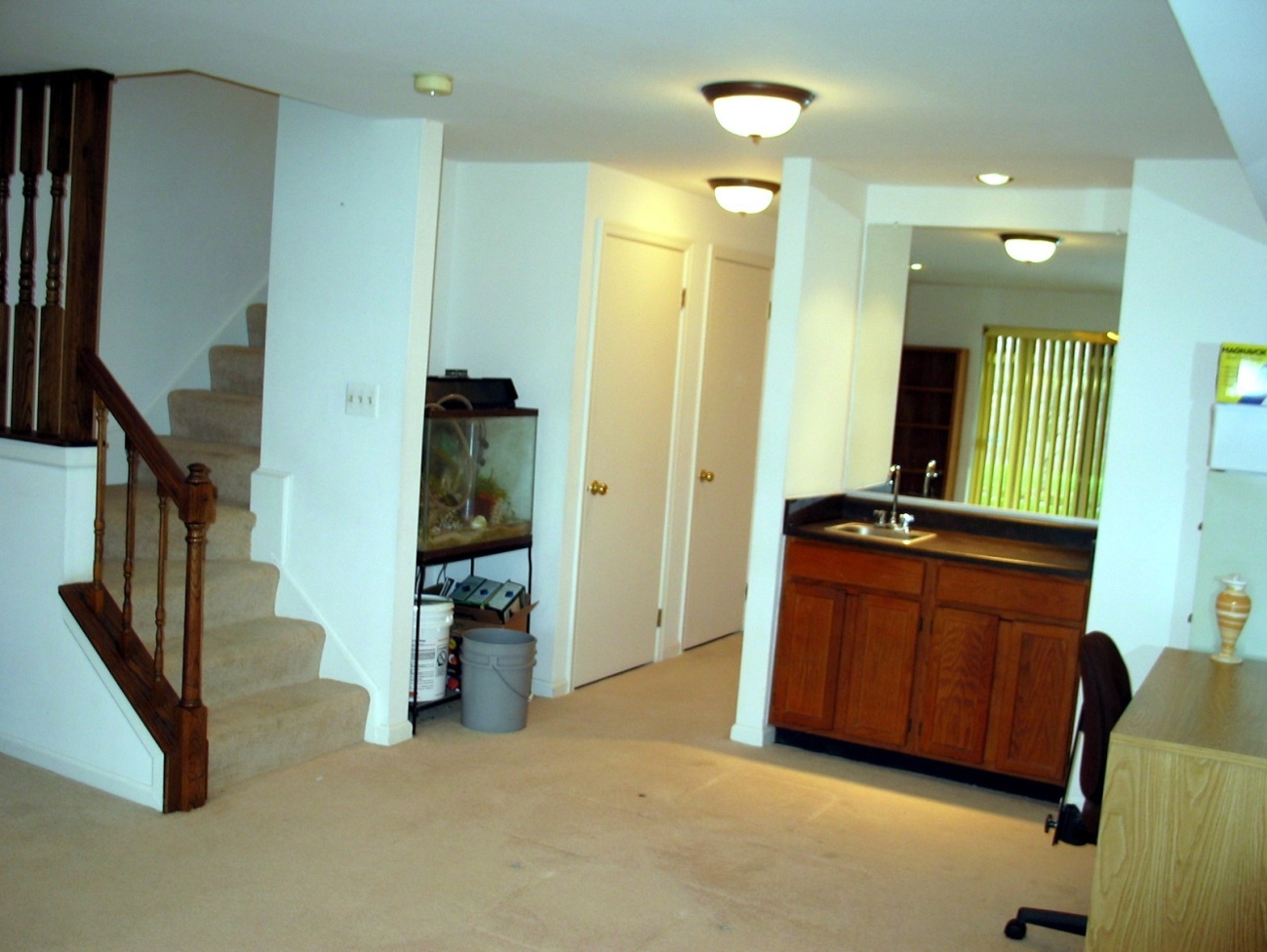}
\end{figure}


After labeling the ceiling first, I wanted to label the ceiling that goes until the end of the scene and then also the curtain and the shelf that we can see at the end. But when I finished labeling all this I realized that the faucet was reflected and then I realized that what I was seeing was a mirror and not a wall on the back. These are typical mistakes that happen when I am zooming into the picture. This is why it is important to go back to the original size from time to time in order to get a good view of the overall scene. 

\begin{figure}[H]
\centering
\includegraphics[width=1\linewidth]{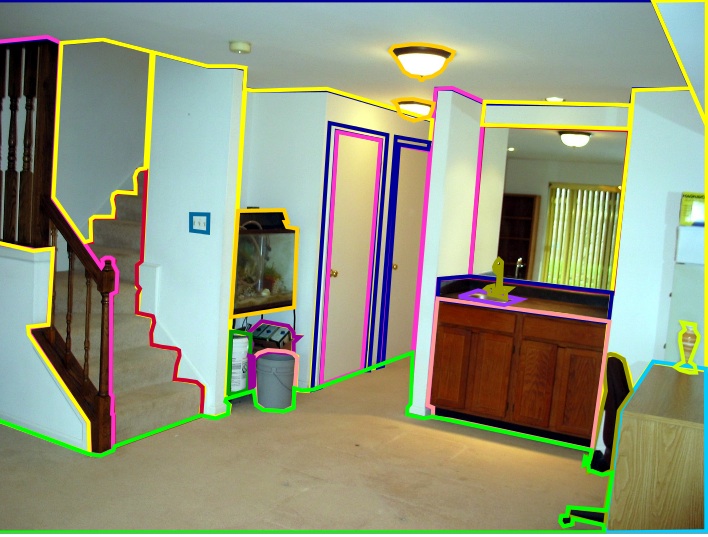}
\end{figure}

\subsection{Object names}

As English is not my mother tongue, I started writing down all the words I was using to make sure that I made no orthographic mistakes. At first, I only wrote down the most frequent words to avoid having to look for them in the dictionary. Soon I had memorized them. But each day the vocabulary was growing and I started organizing them by topics, then I organized them according to the different rooms in a house. I also organized them by subjects, for instance, by all the tools, or all the car appliances. In summary, I was making my own ontologies on the fly with the goal of making the task easier without having to look inside a dictionary and to make consistent annotations by calling all the things always by the same name. 

Each time that I came up with a new organization idea, I had to switch to a new notebook. I never threw the old ones as many times I had to look for a word that was only annotated in one of the old notebooks. 

Everyday I also wrote down the total number of annotated objects to compare it with the final number at the end of the day. This way I had some idea of how many objects I had annotated that day. 

With this job, my memory has improved a lot, almost at the same rate as my pronunciation has gotten worse because I was always making an effort to remember how words were written and I was forgetting how words were pronounced. As a result, I can read in English and I understand almost everything, but I prefer that nobody speaks to me because I do not understand spoken English. 



As I was going from one notebook to the next one, I was also changing the strategy of organizing the object names. 

In the first notebook I started writing all the words in the order that I was using them. One word could be ``banana'' and the next one ``window.'' But as I was progressing in the labeling job this type of dictionary became inefficient whenever I had to look up a particular word. Then I started separating the words into topics but without much emphasis on having a good organization. Therefore, looking up words still remained difficult.


Therefore I decided to make a second notebook where the words were more precisely organized. Unfortunately it was a small notebook and it was not very convenient. Despite this I used it for some time as I got accustomed to it and I knew in which sections I had to look up certain words.


Finally I bought a larger notebook and started classifying the words by topics. This organization is very convenient because I generally work on collections of images that belong to the same topic. My notebook contains topics such as:

\begin{itemize}
\itemsep-0.5em
\item Kitchen
\item Games
\item Bathroom
\item Furniture
\item Music
\item Machines
\item Vegetables
\item Office
\item Car
\item Animals
\item Boats
\item Clothing
\item Chairs and doors
\item Bed accessories
\item Things
\item Airplanes
\item Parks
\item Tools
\end{itemize}


For instance, in the kitchen I can find:
\begin{itemize}
\itemsep-0.5em
\item Frying pan = sarten
\item Pitcher = jarra
\item Mug = jarra de cerveza                                                    
\item dishwasher = lavavajillas
\item Cutting board = tabla                                                        
\item washing machine = lavadora
\item Dish rack = escurreplatos                                                
\item tumble dryer = secadora
\item Knife (set) = cuchillo 
\item Spoon = cuchara
\item Fork = tenedor
\item Saltcellar = salero
\item Bowl = bol
\item Sink = fregadero
\item Strainer = colador 
\item Extractor hood = campana
\item Water cooler = refrigerador de agua
\item Apron = delantal
\item Heater = calentador
\item Refrigerator = frigorifico
\item Blender = licuadora
\item Napkin = servilleta
\item Kettle = tetera para calentar agua
\item Corkscrew = sacacorchos 
\item Pantry = despensa
\item Tureen = sopera
\item Cooker = olla
\item Crock = vasija de barro
\item Skimmer = espumadera
\item Bucket = cubo
\item Place mat = mantel individual 
\item Tongs = pinzas
\item Dishcloth = trapo cocina
\item Worktop = encimera cocina
\item Food processor = robot
\item Utensils canister = bote con cubiertos 
\item Cruet = vinageras 
\item Sauceboat = salsera
\item Mitten = manopla
\item Coffee maker = cafetera
\end{itemize}

Under the topic ``things'' I put objects that I do not know how to classify when I see them. For that reason, sometimes I repeat words without noticing it.

In the ``things'' topic, inside my last notebook, I have:
\begin{itemize}
\itemsep-0.5em
\item Stage = escenario
\item Spotlight = reflector, foco
\item Chinese lantern = farolillo de papel
\item Projection screen = pantalla conferencias
\item Brochures = folletos
\item Pigeonholes = casillero 
\item Label = etiqueta
\item Price tag = etiqueta precio
\item Drawing = dibujo
\item Easel = caballete pintor
\item Trestle = caballete mesa
\item Ceramic figure = figura de ceramica
\item Fireplace utensils = utensilios chimenea
\item Lectern = atril libros
\item Brochure holders = expositor folletos
\item Ashtray = cenicero
\item Folding screen = biombo 
\end{itemize}

There have been objects for which it has been difficult to choose the precise word to use while avoiding using a word already being used for another object. For instance, the spanish word ``biombo'' can be translated as ``screen'' but that word is already being used with a totally different meaning (i.e., computer screen). The same was happening if I used the synonym word ``mampara'' which was also giving me ``screen'' as translation.  I had to specify very clearly the concept I was referring to in order to find different words to express each concept. In the example of the ``screen'' it was not too complex and I decided to use:

\begin{itemize}
\itemsep-0.5em
\item Folding screen = biombo
\item Shower screen = mampara 
\item Movie screen = pantalla
\end{itemize}

And these are pictures of my notebooks that reflect little by little how time passed while working on LabelMe.
\begin{figure}[H]
\includegraphics[width=.43\linewidth]{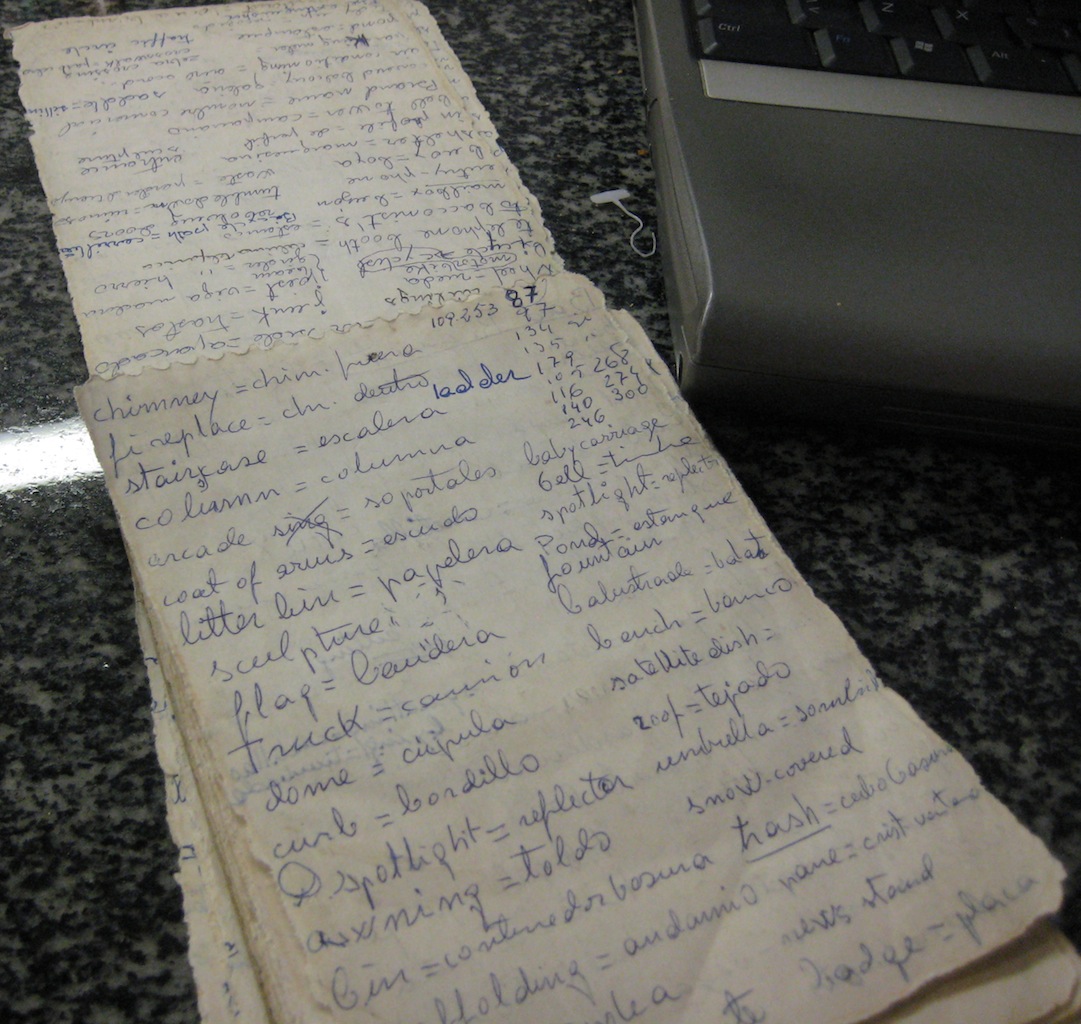}
\includegraphics[width=.54\linewidth]{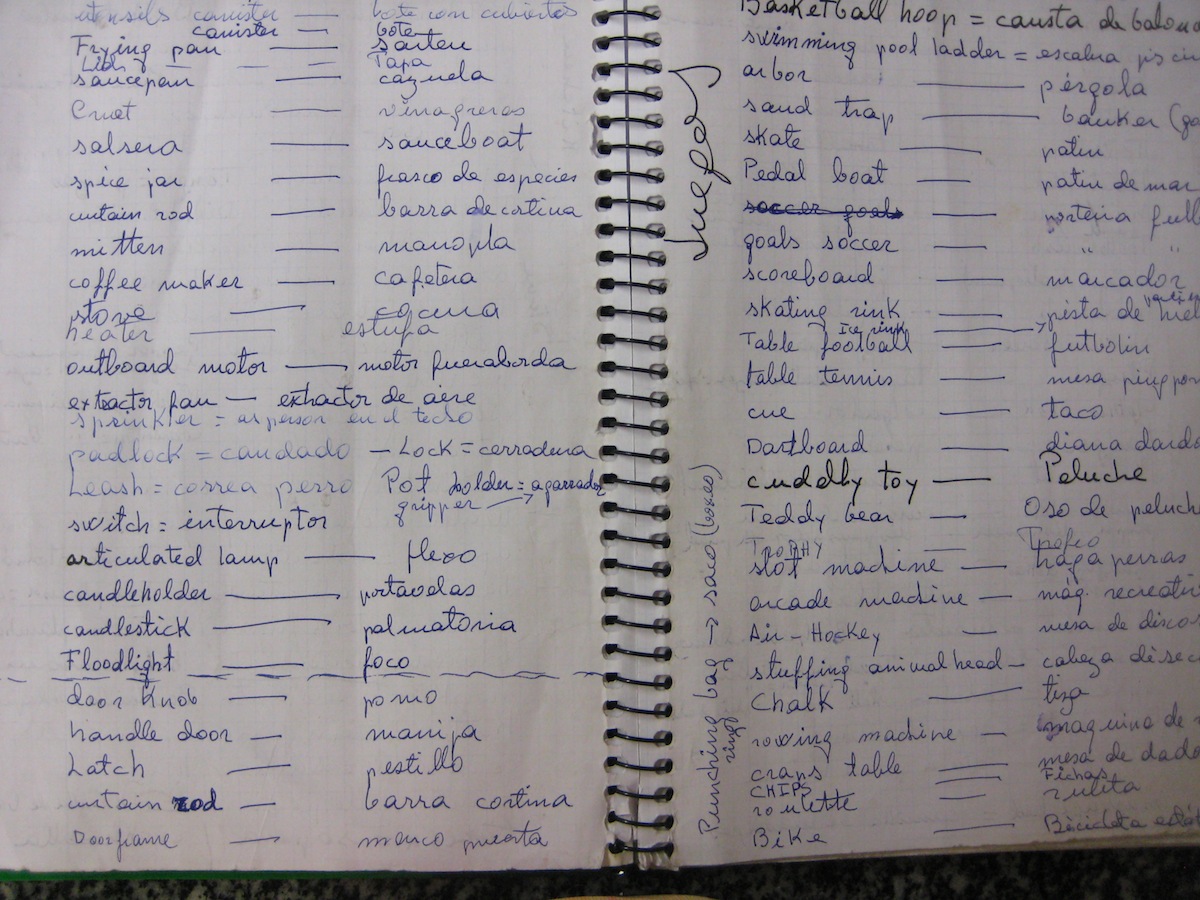}
\end{figure}

\section{Conclusion}

This document might help the reader gain some intuitions about issues that crop up when annotating images. However, this is not an excuse that should prevent the reader from annotating a few images her/himself. Before you ask a computer to segment an image or to recognize an object you should try the task yourself. That will put you one step closer to understanding if you are giving the computer a fair chance to solve the task.

\section{Acknowledgments}
Thanks to Carl Vondrick, Zoya Gavrilov and Aditya Khosla for helping with the editing and translation from spanglish to english. This work is funded by a Google research award, ONR MURI N000141010933 and NSF Career Award No. 0747120.

\bibliographystyle{plain}
\bibliography{biblio}

\end{document}